\documentclass[journal]{elsarticle}

\usepackage{amssymb}
\usepackage{lineno,hyperref}
\usepackage{CJK}
\usepackage{amsmath}
\usepackage{caption}
\usepackage{epstopdf}
\usepackage{wrapfig}
\usepackage{color}
\usepackage{times}
\usepackage{amssymb,amsmath}
\usepackage{chngpage}
\usepackage{array}
\usepackage{graphicx}
\usepackage{float}
\usepackage{subfig}
\usepackage[top=2cm, bottom=2cm, left=2cm, right=2cm]{geometry}
\usepackage{algorithm}
\usepackage{algorithmicx}
\usepackage{algpseudocode}
\usepackage{amsmath}
\usepackage{multirow}
\usepackage{booktabs}
\usepackage{hyperref}

\modulolinenumbers[5]
\usepackage{algorithmicx,algorithm}
\newcommand{\revised}{\textcolor{black}}

\journal{Neural Networks}
\begin{document}
\let\today\relax
\makeatletter
\def\ps@pprintTitle{
    \let\@oddhead\@empty
    \let\@evenhead\@empty
    \def\@oddfoot{\footnotesize\itshape
         {Published by Neural Networks, https://doi.org/10.1016/j.neunet.2022.06.035} \hfill\today}%
    \let\@evenfoot\@oddfoot
    }
\makeatother
\captionsetup{font={normalsize}}
\begin{frontmatter}
\title{Deep Reinforcement Learning Guided Graph Neural Networks \\ for Brain Network Analysis}

\author[IIE,UCAS]{Xusheng Zhao}
\ead{zhaoxusheng@iie.ac.cn}
\author[MAU]{Jia Wu}
\ead{jia.wu@mq.edu.au}
\author[SCST]{Hao Peng\corref{cor1}}
\ead{penghao@buaa.edu.cn}
\author[MAU]{Amin Beheshti}
\ead{amin.beheshti@mq.edu.au}
\author[NAL]{Jessica J.M. Monaghan}
\ead{jessica.monaghan@nal.gov.au}
\author[MAUL]{David McAlpine}
\ead{david.mcalpine@mq.edu.au}
\author[MAUL]{Heivet Hernandez-Perez}
\ead{heivet.hernandez-perez@mq.edu.au}
\author[MAU]{Mark Dras}
\ead{mark.dras@mq.edu.au}
\author[IIE,UCAS]{Qiong Dai\corref{cor1}}
\ead{daiqiong@iie.ac.cn}
\author[NEL]{Yangyang Li}
\ead{liyangyang@cetc.com.cn}
\author[UIC]{Philip S. Yu}
\ead{psyu@uic.edu}
\author[LHU]{Lifang He}
\ead{lih319@lehigh.edu}
\address[IIE]{Institute of Information Engineering, Chinese Academy of Sciences, Beijing, China}
\address[UCAS]{School of Cyber Security, University of Chinese Academy of Sciences, Beijing, China}
\address[MAU]{School of Computing, Macquarie University, Sydney, Australia}
\address[SCST]{Beijing Advanced Innovation Center for Big Data and Brain Computing, Beihang University, Beijing, China}
\address[NAL]{National Acoustic Laboratories, Sydney, Australia}
\address[MAUL]{Department of Linguistics, The Australian Hearing Hub, Macquarie University, Sydney, Australia}
\address[NEL]{National Engineering Laboratory for Risk Perception and Prevention (NEL-RPP), CAEIT, Beijing, China}
\address[UIC]{Department of Computer Science, University of Illinois at Chicago, IL, USA}
\address[LHU]{Computer Science \& Engineering, Lehigh University, PA, USA}
\cortext[cor1]{Corresponding author}

\begin{abstract}
\revised{
Modern neuroimaging techniques enable us to construct human brains as brain networks or connectomes.
}
Capturing brain networks' structural information and hierarchical patterns is essential for understanding brain functions and disease states. Recently, the promising network representation learning capability of graph neural networks (GNNs) has prompted \revised{related} methods for brain network analysis to be proposed. Specifically, these methods apply feature aggregation and global pooling to convert brain network instances into \revised{vector representations encoding brain structure induction} for downstream brain network analysis tasks. However, existing GNN-based methods often neglect that brain networks of different subjects may require various aggregation iterations and use GNN with a fixed number of layers to learn all brain networks. Therefore, how to fully release the potential of GNNs to promote brain network analysis is still non-trivial. 
\revised{
In our work, a novel brain network representation framework, BN-GNN, is proposed to solve this difficulty,} which searches for the optimal GNN architecture for each brain network.
Concretely, BN-GNN employs deep reinforcement learning (DRL) to \revised{automatically predict the optimal number of feature propagations} (reflected in the number of GNN layers) required for a given brain network. \revised{Furthermore, BN-GNN improves the upper bound of traditional GNNs’ performance in eight brain network disease analysis tasks.}
\end{abstract}

\begin{keyword}
Brain Network; Network Representation Learning; Graph Neural Network; Deep Reinforcement Learning.
\end{keyword}
\end{frontmatter}

\section{Introduction} 
\label{Sec_Introduction}
\revised{
With the advancement of modern neuroimaging}, using neuroimaging data effectively has become a research hotspot in both academia and industry~\cite{alexander2007diffusion,huettel2004functional}. Many of these techniques, \revised{e.g.,} diffusion tensor imaging (DTI)~\cite{alexander2007diffusion} and functional magnetic resonance imaging (fMRI)~\cite{huettel2004functional}, enable us to \revised{construct human brains as topological networks (known as ``brain networks'' or ``connectomes'')}~\cite{van2010exploring,urbanski2008brain}. \revised{Unlike brain images, which consist of pixels, nodes/vertexes and edges/links are components of brain networks.
Specifically, nodes in networks usually indicate regions of interest (ROIs), while edges represent connectivity or correlations between ROI pairs.} There are often distinct differences between the brain networks derived from different imaging modalities~\cite{liu2018multi}. For example, DTI-derived brain networks encode the structural connections among ROIs based on white matter fibers, while fMRI-derived brain networks record the functional activity routes of the regions. \revised{Both DTI-derived and fMRI-derived human brain networks are well-researched} and applied to the whole-brain analysis.

The computational analysis of brain networks is becoming more and more popular in healthcare because it can discover meaningful structural information and hierarchical patterns to help understand brain functions and diseases. Here we take brain disease prediction as an example. As one of the most common diseases affecting human health, brain disease has a very high incidence and disability rate, bringing enormous economic and human costs to society~\cite{parisot2018disease}. Considering the complexity and diversity of the predisposing factors of brain diseases, researchers often analyze the brain network states of subjects to assist in inferring the types of brain diseases and then provide reliable and effective prevention or treatment guidelines. For example, \revised{as a geriatric epidemic, Alzheimer's disease (AD) often manifests} as memory and visual-spatial skill impairments~\cite{braak1991neuropathological,zhang2018multi}. Although the existing medical methods cannot effectively treat AD patients, it is possible to delay the onset of AD by tracking the subject's brain network changes and performing interventional therapy in the stage of mild cognitive impairment~\cite{huang2012alzheimer}.

One of the essential techniques in brain network analysis is \revised{brain network representation (a.k.a., network embedding)}~\cite{liu2018multi,cao2017t}, which aims to embed the subjects' brain networks into meaningful low-dimensional representations. These network representations make it easy to separate damaged or special brain networks from normal controls, thereby providing supplementary or supporting information for traditional clinical evaluations and neuropsychological tests.
\revised{As a popular network/graph representation/embedding framework nowadays, graph neural networks (GNNs)~\cite{kipf2016semi,velivckovic2017graph,Hamilton2017inductive} applies convolution to the network structure~\cite{peng2020motif,sun2021sugar}, which learns deep features while reasoning about relational induction within the network~\cite{lecun2015deep,ma2021comprehensive,liu2020deep}.}
Therefore, many GNN-based embedding \revised{algorithms}~\cite{zhang2018multi,arslan2018graph,ktena2018metric} for brain networks have emerged recently. For brain analysis tasks that treat a subject's brain network as an instance, such as brain disease prediction, GNN-based methods first use stackable network modules to aggregate information from neighbors at different hops. This way, they capture brain networks' structural information and hierarchical patterns. More specifically, \revised{GNN learns node-level feature representations by aggregating neighbor information edgewise, where the number of GNN layers controls the total degree of iterative aggregations.}
Then they apply global pooling on the node-level feature matrices to obtain network-level representations.

Although GNN-based \revised{analysis strategies have been successfully introduced into} various brain network analysis tasks, including brain network classification~\cite{arslan2018graph} and clustering~\cite{liu2018multi}, it is still challenging to release the full potential of GNNs on different brain networks. Specifically, existing works usually utilize GNN with a fixed number of layers to learn all brain network instances, ignoring that different brain networks often require distinct optimal aggregation iterations due to structural differences. On the one hand, more aggregation iterations mean considering neighbors at farther hops, which may prompt some brain networks to learn better representations. Unfortunately, increasing the number of aggregations in GNN may also cause over-smoothing problems~\cite{oono2019graph,chen2020measuring}, which means that all nodes in the same network have indistinguishable or meaningless feature representations. On the other hand, it is infeasible to manually specify the number of iterative aggregations for different brain networks, especially when the instance set is large. A straightforward method to alleviate these problems is to deepen the GNN model with skip-connections (a.k.a., shortcut-connections)~\cite{gao2019graph,li2019deepgcns}, which avoid gradient vanishing and a large number of hyperparameter settings. However, this is a sub-optimal strategy because it fails to automate GNN architectures for different brain networks without manual adjustments. 

To solve the above problems, \revised{a novel GNN-based brain network embedding framework (BN-GNN) designed for brain network analysis is developed. With the recent proposal and application} of meta-policy learning~\cite{lai2020policy,zha2019experience}, we expect a meta-policy that automatically determines the optimal number of feature aggregations (reflected in the number of GNN layers) for a given brain network. Specifically, \revised{we heuristically model the optimization and decision iteration of a meta-policy as} a Markov decision process (MDP). First of all, we regard the adjacency matrix of a randomly sampled brain \revised{instance} as the initial state and input it into the policy in MDP. Secondly, we guide the construction of the GNN in MDP based on the action (an integer) corresponding to the maximum Q value output by the policy. Here the action value determines \revised{the stacking of GNN layers, which controls the total count} of feature aggregations on the current brain network. Thirdly, we pool the node features into a network representation. Then we perform network classification to optimize the current GNN and employ a novel strategy to calculate the current immediate reward. Fourthly, we sample the next network instance through a heuristic state transition strategy and record the state-action-reward-state quadruple of this process. In particular, we apply the double deep q-network (DDQN)~\cite{mnih2015human,van2016deep}, a classic deep reinforcement learning (DRL)~\cite{arulkumaran2017deep} algorithm, to simulate and optimize the policy. Finally, we utilize the trained policy (i.e., meta-policy) as meta-knowledge to guide the construction and training of another GNN and implement specific brain network analysis tasks.

Overall, \revised{the main results are condensed as:}

$\bullet$ A novel \revised{brain} network representation learning framework (i.e., BN-GNN\footnote{https://github.com/RingBDStack/BNGNN}) through GNN and DRL is proposed to assist brain network analysis \revised{tasks. In this way, the number of GNN feature aggregations can be altered} for different brain networks, thereby releasing the full potential of traditional GNNs in brain network representation learning. 

$\bullet$ \revised{This is the first study in the field of brain network analysis to introduce DRL into the GNN model.} We are also the first to use GNNs with different layers to learn different subjects' brain networks.

$\bullet$ \revised{Experiments on eight brain network disease analysis datasets (e.g., BP-DTI and HA-EEG) show that BN-GNN stands out among many advanced algorithms, improving the upper bound of traditional GNNs’performance.}

The remainder of \revised{the article is briefly described as:} Sec.~\ref{Sec_Relatedwork} and Sec.~\ref{Sec_Preliminaries} describe related work and preliminary knowledge, respectively. Sec.~\ref{Sec_Methodology} details the implementation of BN-GNN \revised{with} graph convolutional network (GCN)~\cite{kipf2016semi} and DDQN~\cite{van2016deep}. Sec.~\ref{Sec_Experiments} \revised{gives} the experimental results and corresponding analysis. Sec.~\ref{Sec_Conclusions} summarizes \revised{our work}.

\section{Related Work} 
\label{Sec_Relatedwork}
\subsection{Graph Neural Networks}
\revised{GNNs are currently the preferred strategy for processing topological graph data, which follows the principle that neighbor information affects the feature embedding of central nodes edgewise. Spectral- and spatial-based are two major labels of existing GNN models. A representative product of GNN models is GCN~\cite{kipf2016semi},} which is inspired by the traditional convolutional \revised{operations} on images in the Euclidean space. \revised{To widen and improve aggregation performance bounds and explainability, the graph attention network (GAT)~\cite{velivckovic2017graph} refines the relative importance of neighbors from the perspective of target nodes.} To improve the aggregation efficiency of large-size graphs, GraphSAGE~\cite{Hamilton2017inductive} \revised{extracts a consistent and predefined number of local neighbors for each target one.} In addition, many GNN variants based on the above methods or frameworks have been proposed, and they have made outstanding contributions, such as in clinical medicine~\cite{zhang2018multi}.

\subsection{Reinforcement Learning Guided Graph Neural Networks}
Recently, with the advances of \revised{RL}, many works combine RL with GNNs to further raise the performance boundary of GNNs. For example,~\cite{dou2020enhancing} proposed a GNN \revised{algorithm, i.e.,} CARE-GNN to improve the capability to recognize fraudsters in fraud inspection tasks. 
CARE-GNN first sorts the neighbors based on their credibility and then uses RL to guide the traditional GNN to filter out the most valuable neighbors for each node to avoid fraudsters from interfering with normal users.~\cite{peng2021reinforced} proposed a novel recursive and reinforced graph neural network framework learn more discriminative and efficient node representation from multi-relational graph data.~\cite{Peng2022reinforced} proposed \revised{a GNN that considers different relations to achieve network representation, which is based on multi-agent RL for relation importance assignment.} ~\cite{gao2019graphnas} proposed a graph \revised{NAS} algorithm, namely GraphNAS. GraphNAS first utilizes a recurrent network to create variable-length strings representing the architectures of GNNs and then applies RL to update the recurrent network for maximizing the \revised{quality of model building.}~\cite{nishi2018traffic} proposed a GCN-based algorithm for traffic signal control called NFQI, which applies a model-free RL approach to learn responsive traffic control in order to deal with temporary traffic demand changes when environmental knowledge is insufficient.

Since deep RL (DRL) combines the perception capability of deep learning with the decision capability of RL, it is a new research hotspot \revised{in} artificial intelligence. For example,~\cite{yan2020automatic} proposed a virtual network embedding algorithm (i.e., V3C+GCN) that combines DRL with a GCN-based module.~\cite{lai2020policy} proposed a meta-policy framework (i.e., Policy-GNN), which adaptively learns an aggregation strategy to use DRL to perform various aggregation iterations on different nodes. Though the above methods directly or indirectly use RL or DRL to improve GNNs, there is still no work using DRL to guide GNNs to assist brain network analysis, which often requires different models for different brain networks.

\subsection{GNN-based Brain Network Representation Learning}
Unlike traditional shallow methods for brain network representation learning, such as tensor decomposition~\cite{liu2018multi,cao2017t}, some works use GNNs to capture deep feature representations of brain networks for downstream brain analysis tasks. Concretely,~\cite{li2020pooling} proposed a PR-GNN that includes regularized pooling layers, which calculates node pooling scores to infer which brain regions are obligatory parts of certain brain disorders.~\cite{bi2020gnea} \revised{proposed an aggregator that applies extreme learning machines (ELMs), which avoids tuning iterations and widens the feature passing performance boundaries. In addition, they provided a GNEA model based on the aforementioned aggregator to enable brain graph analysis. Hi-GCN method proposed by ~\cite{jiang2020hi} can perform hierarchical embedding of brain networks. In order to improve the accuracy of brain disease analysis, Hi-GCN considers graph structure induction and also introduces patient group-level structural information.~\cite{ma2019deep} developed an graph learning algorithm for brain network analysis, namely HS-GCN, which utilizes two GCNs to build a siamese model and learns brain network representations by means of supervised metrics.}~\cite{zhong2020eeg} proposed a regularized GNN (i.e., RGNN) for emotion recognition based on electroencephalogram.~\cite{xing2021ds} \revised{developed a GCN-based algorithm (i.e., DS-GCNs) that can condense meaningful representations from functional connections readily available in neuroanalytical tasks.} DS-GCNs calculate a dynamic functional connectivity matrix with a sliding window and implement a long and short-term memory layer based on graph convolution to process dynamic graphs.

\revised{Corresponding to unimodal brain connectome studies, GNNs are also quite popular in multi-modality brain analysis scenarios.} For example,~\cite{zhang2018multi} \revised{proposed a GCN model (i.e., MVGCN) for combining different view information in brain analysis tasks, helping to distinguish Parkinson's disease cases from healthy controls.~\cite{gurbuz2021mgn} proposed a multi-view normalization network based on GNN (i.e., MGN-Net), which normalizes and combines a set of multi-view brain networks into one.}

Although these GNN-based single- or multi-modality methods have made significant breakthroughs in many brain network analysis tasks, they have failed to implement customized aggregation for different subjects' brain networks in experiments such as brain disease prediction.

\section{Preliminaries} 
\label{Sec_Preliminaries}
\revised{First of all, we formulate the brain network analysis. Next, we introduce the network representation learning method when the GNN layer is predefined, MDP, and DRL. The key notations/symbols are given in Table~\ref{tab_Notation}.} 

\subsection{Problem Formulation}
\label{Sec_Problem1}
Generally, a brain \revised{connectome can be abstracted as a graph $G=(V, E)$, where $V = \begin{Bmatrix}v_{1},\dots,v_{n}\end{Bmatrix}$ indicates the node set, $E$ contains weighted edges that represent topological relations among nodes. Let $\mathbf{W}$ denote $G$'s initial weighted matrix, so $\mathbf{W}(i, j)$ means the edge correlation between $v_i$ and $v_j$ (which may tend to zero for no or weak connection).} Let $D=\begin{Bmatrix}G_{1},\dots,G_{m}\end{Bmatrix}$ \revised{be an ensemble of brain networks based on $m$ brain subjects.} We assume \revised{that these network instances have different structures but the same nodes}, where a specific region division strategy determines the number of nodes. Given the $k$-th brain network $G_{k}=(V_{k},E_{k})$, we abstract it as a weighted matrix $\mathbf{W}_{k}\in\mathbb{R}^{n\times n}$.

\revised{We focus on the problem of brain network representation learning based on DRL-introduced GNNs,} which is used for both classification and clustering. Specifically, we focus on the classification task since it is \revised{often the research basis for brain network analysis.} Given dataset $D$, we assume that the corresponding network labels $\mathbf{Y}$ are known. For convenience, we denote the training, validation, and test set of $D$ as $D_{train}$, $D_{val}$, and $D_{test}$, respectively, where $D=D_{train}\cup D_{val} \cup D_{test}$. Based on the brain networks in $D_{train} \cup D_{val}$, we first continuously optimize a policy $\pi$. Next, we employ the trained policy (i.e., meta-policy) to guide the construction of GNN and utilize the customized GNN to learn the node representations of each brain network that meet the number of feature aggregations. Then, by applying the global pooling at the last layer of GNN, we convert the node-level feature tensor into a low-dimensional network-level representation matrix $\mathbf{E}$, allowing brain network instances with different labels to be easily separated. Last, we feed $\mathbf{E}$ into the full-connected layer to perform brain network classification. 

\begin{table}[t]
\small
\caption{Glossary of Notations.}
\vspace{-1mm}
\centering
\begin{tabular}{lllll}
\hline
Notation & \multicolumn{4}{l}{Definition}   \\ \hline
$D$; $D_{train}$; $D_{val}$; $D_{test}$ & \multicolumn{4}{l}{The brain network dataset; The training set of $D$; The validation set of $D$; The test set of $D$}   \\
$G$; $V$; $E$ & \multicolumn{4}{l}{The brain network; \revised{Nodes in $G$; Edges in $G$}}   \\
$S; A$ & \multicolumn{4}{l}{The state space in MDP; The action space in MDP}   \\
$\mathbf{W}; \mathbf{A}; \widehat{\mathbf{A}}$ & \multicolumn{4}{l}{The initial weighted matrix of $G$; The adjacency matrix of $G$; The normalized form of $\widetilde{\mathbf{A}}$}   \\
$\widetilde{\mathbf{A}}; \widetilde{\mathbf{D}}$ & \multicolumn{4}{l}{\revised{The self-loop form of $\mathbf{A}$}; The degree matrix of $\widetilde{\mathbf{A}}$}   \\
$\mathbf{E}; \mathbf{F}$ & \multicolumn{4}{l}{The network-level feature matrix of $D$; The node-level feature matrix of $G$}   \\
$\mathbf{T}$ & \multicolumn{4}{l}{The feature transformation \revised{operator}} \\
$\mathbf{C}; \widehat{\mathbf{C}}$ & \multicolumn{4}{l}{The importance coefficient matrix of $G$; The normalized form of $\mathbf{C}$}   \\
$m; n$ & \multicolumn{4}{l}{\revised{Total count of brain networks of $D$; Total node count of $V$}}   \\ 
$d$ & \multicolumn{4}{l}{The \revised{vector} representation dimension of $\mathbf{E}$ or $\mathbf{F}$}   \\
$l$ & \multicolumn{4}{l}{The total number of \revised{GNN layers or iterative aggregations}}   \\
$i; j; k$ & \multicolumn{4}{l}{These notations represent index variables}   \\
$s; a; r$ & \multicolumn{4}{l}{The state in MDP; The action in MDP; The reword in MDP}   \\
$t; b$ & \multicolumn{4}{l}{\revised{The final timestep value of MDP; Set size containing all actions}} \\
$w$ & \multicolumn{4}{l}{The window size of the history records in $REW(\cdot)$} \\
$\oplus$ & \multicolumn{4}{l}{The feature combination operation, such as summation and concatenation}   \\
$\pi$ & \multicolumn{4}{l}{The policy function in MDP or the meta-policy}   \\
$\gamma; \epsilon$ & \multicolumn{4}{l}{The discount coefficient of $\mathcal{R}$; The epsilon probability of exploration of $\pi$}   \\
$\sigma(\cdot)$ & \multicolumn{4}{l}{The activation function, such as $Tanh$ and $ReLU$}   \\
$AGG(\cdot)$ & \multicolumn{4}{l}{The feature aggregation function of GNN, such as convolution and attention}   \\ 
$REW(\cdot)$ & \multicolumn{4}{l}{The immediate reward function in MDP}   \\
$PER(\cdot)$ & \multicolumn{4}{l}{The classification performance metric, such as accuracy.}   \\
$\mathcal{R}$ & \multicolumn{4}{l}{The discounted cumulative return in MDP}   \\
$\mathcal{Q}_{eval}; \mathcal{Q}_{target}$ & \multicolumn{4}{l}{The evaluation DNN in DDQN; The target DNN in DDQN}   \\
$\mathcal{L}_{GNN}; \mathcal{L}_{policy}$ & \multicolumn{4}{l}{The training loss of GNN; The training loss of meta-policy}   \\
\hline
\end{tabular}
\label{tab_Notation}
\vspace{-1mm}
\end{table}

\subsection{Learning Network Representations with Layer-fixed GNN}
\label{Sec_Problem2}
GNNs learn node-level feature representations through the network structure. \revised{Given an instance $G=(V,E)$, we receive} its adjacency matrix $\mathbf{A}\in\mathbb{R}^{n\times n}$ and initial \revised{region features} $\mathbf{F}^{(0)}\in\mathbb{R}^{n \times d^{(0)}}$, \revised{and then} express the feature aggregation process of \revised{$v_{i}\in V$ in a layer-fixed GNN} as follows~\cite{dou2020enhancing}:
\begin{equation}
    \small
	\begin{split}
	\mathbf{F}^{(l)}(i) &= \sigma
	\begin{pmatrix} 
	\mathbf{F}^{(l-1)}(i)
	\oplus
	AGG^{(l)}
	\begin{pmatrix}
	\begin{Bmatrix}
	\mathbf{F}^{(l-1)}(j):\mathbf{A}(i,j)>0 \end{Bmatrix}
	\end{pmatrix}\end{pmatrix},
	\end{split}
	\label{eq_AGG}
\end{equation}
where $\mathbf{F}^{(l-1)}\in\mathbb{R}^{n\times d^{(l-1)}}$ and $\mathbf{F}^{(l)}\in\mathbb{R}^{n\times d^{(l)}}$ \revised{indicate the input and output features of the model. $AGG^{(l)}$ indicates the aggregation module, their superscript $(l)$ indicates that the features or modules belong to the $l$-th layer}. $\oplus$ is an operation used to \revised{fuse the features of $v_{i}$ and its neighbors.} $\sigma$ \revised{means the activation function like $Tanh$}. It is worth noting that $\mathbf{A}$ should be reliable as it remains unchanged at all layers. Taking graph convolutional network (GCN)~\cite{kipf2016semi} with two \revised{aggregations} as an example, it implements Eq.~\ref{eq_AGG} through convolution:
\begin{equation}
    \small
	\begin{split}
	&\widehat{\mathbf{A}}=\widetilde{\mathbf{D}}^{-1/2}\widetilde{\mathbf{A}}\widetilde{\mathbf{D}}^{-1/2}\\
	&\mathbf{F}^{(2)}=ReLU(\widehat{\mathbf{A}}ReLU(\widehat{\mathbf{A}}\mathbf{F}^{(0)}\mathbf{T}^{(1)})\mathbf{T}^{(2)})
	\end{split},
	\label{eq_GCN}
\end{equation}
where $\widehat{\mathbf{A}}\in\mathbb{R}^{n\times n}$ is the symmetrical normalized form of $\widetilde{\mathbf{A}}$, $\widetilde{\mathbf{A}}=\mathbf{A}+\mathbf{I}$ is the adjacency matrix \revised{with the identity matrix $\mathbf{I}\in\mathbb{R}^{n\times n}$ added}, and $\widetilde{\mathbf{D}}\in\mathbb{R}^{n\times n}$ is the degree matrix of $\widetilde{\mathbf{A}}$. At the first layer, since $\widehat{\mathbf{A}}$ encodes each node's direct (1-hop) neighbor information, $\widehat{\mathbf{A}}\mathbf{F}^{(0)}$ essentially implements the first convolution aggregation through summation. At the second layer, the continuous multiplication of the adjacency matrix (i.e., $\widehat{\mathbf{A}}ReLU(\widehat{\mathbf{A}})$) makes the neighbor's neighbor (2-hop) information included in the second aggregation. Therefore, \revised{when GNN models stack and aggregate more}, the receptive field of GNN becomes wider, so more neighbors participate in aggregation. In addition, $\mathbf{T}^{(1)}\in\mathbb{R}^{d^{(0)}\times d^{(1)}}$ and $\mathbf{T}^{(2)}\in\mathbb{R}^{d^{(1)}\times d^{(2)}}$ are the learnable matrices for feature transformation of the first and second layer, respectively. Unlike GCN, which equally distributes the importance of all neighbors, \revised{GAT~\cite{velivckovic2017graph} counts neighborhood importance weights through attention when summing their features.} Taking the first layer of a single-head GAT as an example, the node feature aggregation \revised{is} as follow:
\begin{equation}
    \small
	\begin{split}
	&\mathbf{C}(i,j)=(\mathbf{F}^{(0)}(i)\mathbf{T}^{(1)}\oplus \mathbf{F}^{(0)}(j)\mathbf{T}^{(1)})q^{T}\\
	&\widehat{\mathbf{C}}=softmax(\mathbf{C}(i,j))=\frac{exp(ReLU(\mathbf{C}(i,j)))}{\sum_{v_{k}\in V(i)} exp(ReLU(\mathbf{C}(i,k)))}\\
	&\mathbf{F}^{(1)}(i)=ReLU(\sum_{v_{j}\in V(i)}\widehat{\mathbf{C}}\mathbf{F}^{(0)}(j)\mathbf{T}^{(1)})
	\end{split},
	\label{eq_GAT}
\end{equation}
where $\mathbf{F}^{(0)}(i)\in\mathbb{R}^{1\times d^{(0)}}$ represents the initial feature representation of node $v_{i}$, $\mathbf{T}^{(1)}\in\mathbb{R}^{d^{(0)}\times d^{(1)}}$ is the feature transformation matrix whose parameters are shared, $\oplus$ is the concatenation operation, and $q\in\mathbb{R}^{1\times 2d^{(1)}}$ is the attention feature vector. $\mathbf{C}(i,j)$ is the importance coefficient (a real number) of node $v_{j}$ to node $v_{i}$, $\widehat{\mathbf{C}}$ is the normalized form of $\mathbf{C}$, and $\mathbf{F}^{(1)}(i)\in\mathbb{R}^{1\times d^{(1)}}$ is the \revised{transformed representation vector of the target $v_{i}$.} Here $V(i)=\begin{Bmatrix}v_{j}:\mathbf{A}(i,j)>0\end{Bmatrix}$ indicates the neighbor set of node $v_{i}$. Similarly, GAT also controls the number of feature aggregations by changing the number of layers. After the final aggregation is completed at the last layer $l$, GNN-based methods perform global pooling on all nodes to obtain the final network representation. The process of global average pooling is \revised{described below}:
\begin{equation}
    \small
	\begin{split}
	\mathbf{E}(i) = \frac{1}{n}\sum_{v_{j}\in V} \mathbf{F}^{(l)}_{i}(j),
	\end{split}
	\label{eq_Pool}
\end{equation}
where $\mathbf{E}(i)\in\mathbb{R}^{1\times d^{(l)}}$ and $\mathbf{F}^{(l)}_{i}(j)$ are the network-level feature vector and node-level feature matrix of the $i$-th network instance $G_{i}$, respectively. Then the cross entropy loss of this part \revised{is} as follows:
\begin{equation}
    \small
	\begin{split}
	\mathcal{L}_{GNN}=-\sum_{G_{i}\in D_{train}} log(\mathbf{E}(i)\mathbf{T}^{(l+1)})\mathbf{Y}(i)^{T},
	\end{split}
	\label{eq_GNN_Loss}
\end{equation}
where $\mathbf{T}^{(l+1)}$ is the fully connected layer \revised{applied as a classifier, $\mathbf{Y}(i)$ indicates the $i$-th brain instance's class label.}

\revised{In brain network disease analysis, most instances} can be abstracted as weighted matrices describing the connections among brain regions, but they often do not have initial \revised{region} features. \revised{Besides, there is little research on constructing informative node features and edges for GNN-based brain network learning.} A common strategy is to use the initial weighted matrix $\mathbf{W}$ associated with each brain network $G$ as its initial node features (i.e., $\mathbf{F}^{(0)} = \mathbf{W}$) and define a group-level adjacency matrix $\mathbf{A}$ for GNN. For example,~\cite{zhang2018multi} defined $\mathbf{A}$ as a \revised{coarse-grained network processed by k-nearest neighbor (KNN)}, and~\cite{zhang2019new} \revised{constructed the Laplacian operator through the small-world model to infer $\mathbf{A}$ in the process of representation learning.} However, using the same adjacency matrix for different brain networks may blur the differences between different networks. Different from the previous work~\cite{zhang2018multi,zhang2019new}, our goal is to generate a  separate adjacency matrix for each brain network and implement network representation learning for different brain networks based on GNNs with different layers.

\subsection{Markov Decision Process}
\label{Sec_Problem3}
\revised{MDP is a natural description of sequential decision problems}, used to simulate the random actions and rewards that the agent can achieve in an environment with Markov properties. Here we denote the MDP as a quintuple $(S,A,\pi,REW,\mathcal{R})$, where \revised{$S$ and $A$ mean the state and action set/space}, $\pi$ is the policy that outputs \revised{the action conditional probability distribution of the input state}, $REW:S\times A\rightarrow \mathbb{R}$ is the immediate reward function, and $\mathcal{R}$ is the accumulation of rewards over time (a.k.a, return). The decision process in each timestep $i\in[1,t]$ is as follows: the agent first perceives the current state $s_{i}\in S$ and \revised{then implement an action as directed by} $\pi$. Then, the environment (which is affected by the action $a_{i}$) feeds back to the agent the next state $s_{i+1}$ as well as a reward $r_{i}=REW(s_{i},a_{i})$. In the standard MDP, our goal is to train \revised{$\pi$ to maximize the accumulation of discounted returns. The total return can be expressed in summed form as:}
\begin{equation}
    \small
	\begin{split}
	\mathcal{R} = r_{1} + \gamma r_{2} + \gamma^{2} r_{3} + \dots + \gamma^{t-1} r_{t} =\sum_{i=1}^{t}\gamma^{i-1}r_{i},
	\end{split}
	\label{eq_MDP}
\end{equation}
where $\gamma\in(0,1)$ is the discount coefficient \revised{to constrain future rewards with low reliability. The optimization steps for policy $\pi$ are detailed in the next section.}

\begin{figure}[t]
    \centering
    \includegraphics[width=1.\textwidth]{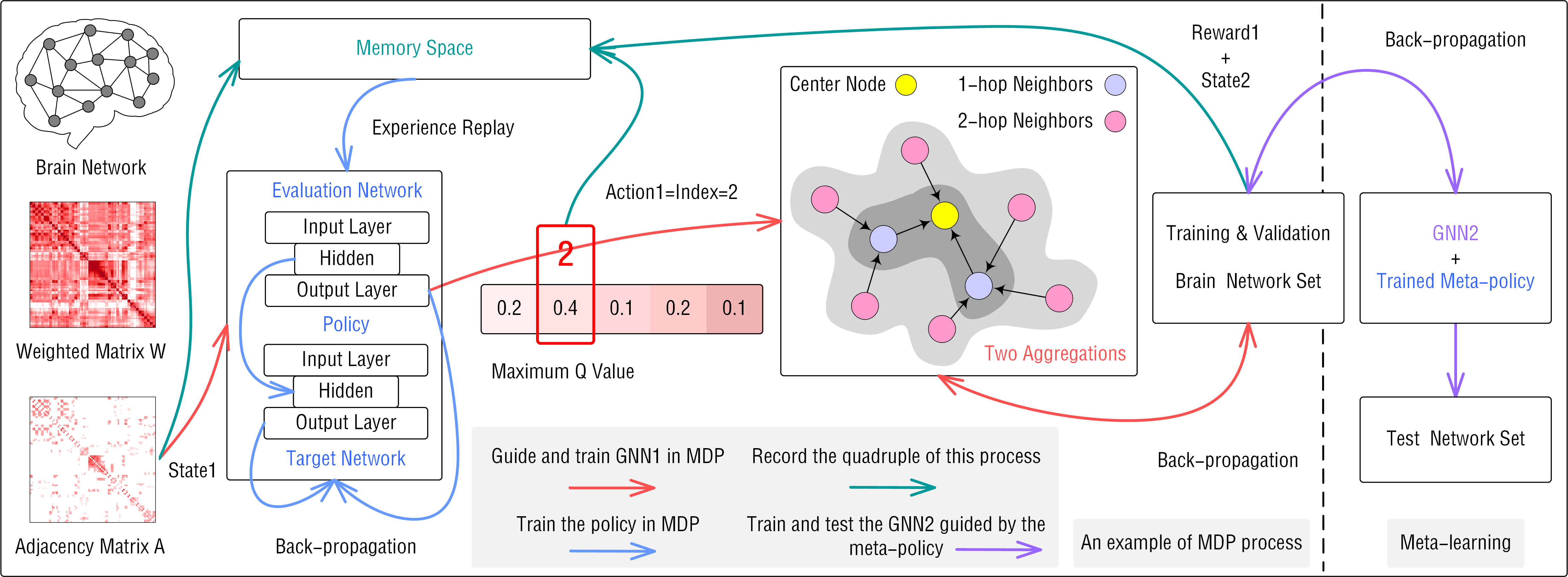}
    \caption{\revised{Illustration of our} framework BN-GNN for brain network representation learning. The left side of the dotted line illustrates an example of the MDP process. \revised{First of all, we treat the adjacency matrix derived from the brain network} building module as the current state and input it into the policy. \revised{Then we follow the guiding signals (Q values) output by the policy to decide the current action.} Here we treat the index of the selected action in the action space as the action value to guide the number of aggregations of the current brain network in GNN. After feature aggregation, \revised{we employ global pooling to harvest network instance-level representations and train GNN models in MDP.} Subsequently, the current reward is calculated by comparing the performance changes on the verification set. \revised{Finally, the transition strategy is applied to obtain the state of the next timestep.} To train the policy in MDP, we record the state-action-reward-state quadruple of this process to the memory space and follow the DDQN method to calculate the loss of the policy. On the right side of the dotted line, we apply the trained meta-policy to guide the training of a new GNN and perform brain network analysis tasks.
    } 
    \label{fig_Framework}
\end{figure}

\subsection{Solving MDP with Deep Reinforcement Learning}
\label{Sec_Problem4}
In many scenarios, the state space $S$ is huge or inexhaustible. 
In this case, it is sub-optimal or infeasible to train the policy $\pi$ by maintaining and updating a state-action table. Deep reinforcement learning (DRL)~\cite{arulkumaran2017deep} is an effective solution because it can use neural networks to simulate and approximate the actual relationship between any state and all possible actions. Here, we focus on a classic DRL algorithm called double deep q-learning (DDQN)~\cite{mnih2015human,van2016deep}, which uses two deep neural networks (DNNs) to simulate the policy $\pi$. Concretely, in each timestep $i$, DDQN first inputs the current state $s_{i}$ into the evaluation DNN to \revised{get the predicted values} of all actions and regard the action corresponding to the maximum Q value as the current action, which should conform to the following formula:
\begin{equation}
    \small
	\begin{split}
	a_{i} = \begin{cases}
	random\: action, &\text{w.p.}\:\epsilon\\
	argmax_{a_{i}}(Q_{eval}(s_{i},a_{i})), &\text{w.p.} \:1-\epsilon\end{cases},
	\end{split}
	\label{eq_Action}
\end{equation}
where $\epsilon$-greedy \revised{can make DDQN more portable} and avoid the dilemma of exploration and utilization. The maximum Q value $max_{a_{i}}(Q(s_{i},a_{i}))$ is essentially the expected maximum discounted return in the current state $s_{i}$, and the corresponding Bellman equation can be expressed as follows:
\begin{equation}
    \small
	\begin{split}
	max_{a_{i}}(Q_{eval}(s_{i},a_{i}))=max(\mathcal{R}_{i})&=max(r_{i}+\gamma(r_{i+1}+\gamma(r_{i+2}+\dots)))\\
	&=r_{i}+\gamma max(\mathcal{R}_{i+1}).
	\end{split}
	\label{eq_Q}
\end{equation}
After determining the current action $a_{i}$, DDQN uses the reward function (designed according to the specific environment) to calculate the current actual reward (i.e., $r_{i}=REW(s_{i},a_{i})$) and then performs state transition to obtain the next state $s_{i+1}$. Moreover, there is a memory space in DDQN that records each MDP process (also known as ``experience'' and recorded as a state-action-reward-state quadruple $<s_{i},a_{i},r_{i},s_{i+1}>$). To optimize DNNs with experience replay, DDQN first records the current experience and then randomly extracts a memory block from the memory space. For example, through the experience $<s_{i},a_{i},r_{i},s_{i+1}>$, the loss of DNNs (i.e., policy) can be calculated as follows:
\begin{equation}
    \small
	\begin{split}
	\mathcal{L}_{policy}&=(max(\mathcal{Q}_{target}(s_{i},a_{i}))-max(\mathcal{Q}_{eval}(s_{i},a_{i})))^{2}\\
	&=(r_{i}+\gamma max_{a_{i+1}}(\mathcal{Q}_{target}(s_{i+1},a_{i+1}))-max_{a_{i}}(\mathcal{Q}_{eval}(s_{i},a_{i})))^{2},
	\end{split}
	\label{eq_Policy_Loss}
\end{equation}
where state $s_{i}$ is input to the evaluation DNN to obtain the predicted maximum Q value $max_{a_{i}}(\mathcal{Q}_{eval}(s_{i},a_{i}))$ in the $i$-th timestep, the next state $s_{i+1}$ is input to the target DNN to calculate the maximum Q value $max_{a_{i+1}}(\mathcal{Q}_{target}(s_{i+1},a_{i+1})$ in the next timestep, and $r_{i}$ is the actual reword. Based on the Bellman equation, DDQN takes the $max(\mathcal{Q}_{target}(s_{i},a_{i}))$ output from the target DNN as the actual maximum Q value in timestep $i$ and trains the evaluation DNN through the back-propagation algorithm. It is worth noting that DDQN does not update the target network through loss but copies the parameters of the evaluation DNN to the target DNN. In this way, DDQN effectively alleviates the over-estimation problem often found in DRL. Since DDQN uses two DNNs to simulate the policy $\pi$, the above is also the training process of $\pi$.

\section{Methodology} 
\label{Sec_Methodology}
Fig.~\ref{fig_Framework} \revised{illustrates the brain network representation learning framework BN-GNN}, which consists of three modules: network building module, meta-policy module and GNN module. The network building module provides the state space for the meta-policy module. The \revised{meta one utilizes feedbacks} (i.e., rewards) of the GNN module to search for the optimal meta-policy $\pi$ continuously, and the GNN module performs brain network representation learning according to the guidance (i.e., actions) of the meta-policy. Next, we introduce the technical details of each module.

\subsection{Network Building Module}
\label{Sec_Net_Module}
The network building module generates adjacency matrices from the brain networks‘ initial weighted matrices, providing state space for the meta-policy module. In GNN, \revised{feature aggregation that relies on adjacency matrix $\mathbf{A}$ is essential. Therefore, $\mathbf{A}$} should be appropriately designed to reflect the neighborhood correlations because it directly affects the node feature representation learning. Inspired by the previous work in~\cite{zhang2018multi}, \revised{we utilize KNN to construct reliable adjacency matrices to advance learning brain network representation for GNNs. Specifically, taking a brain instance} $G=(V,E)$, we utilize its weighted matrix $\mathbf{W}=\mathbf{F}^{(0)}$ and KNN to obtain reliable neighbors $V(i)$ of any node $v_{i} \in V$. If $v_{i}\in V(j)$ or $v_{j}\in V(i)$, then $\mathbf{A}(i,j) = 1$ and $\mathbf{A}(j,i) = 1$, otherwise \revised{both are zero}. After that, we calculate new edge confidences to refine the reliable \revised{matrix $\mathbf{A}$}:
\begin{equation}
	\begin{split}
	&\mathbf{A}(i,j)=\begin{cases}exp(-\left\|\mathbf{F}^{(0)}(i)-\mathbf{F}^{(0)}(j)\right\|), &\mathbf{A}(i,j)=1\\
	0,&\mathbf{A}(i,j)=0
	\end{cases},
	\end{split}
	\label{eq_NetBuild}
\end{equation}
where $\mathbf{F}^{(0)}(i)$ is \revised{$\mathbf{F}^{(0)}$'s $i$-th row and $v_{i}$'s initial feature embedding. The module} is also used to construct the subject network in the state transition strategy, which will be introduced in the following subsection.

\subsection{Meta-policy Module}
\label{Sec_Policy_Module}
The meta-policy module trains a policy that can be viewed as meta-knowledge to determine the number of aggregations of brain network features in GNN. As mentioned in Sec.~\ref{Sec_Problem3}, \revised{the learning of the policy $\pi$ is abstracted as an MDP} that contains five essential components, i.e., $(S, A, \pi, REW, \mathcal{R})$. Here, we give the relevant definitions in the context of brain network embedding in timestep $i$.

$\bullet$ State space ($S$): The state $s_{i}\in S$ represents the adjacency matrix of the brain \revised{instance}.

$\bullet$ Action space ($A$): The action $a_{i}\in A$ determines the number of iterations for feature aggregation that the brain network requires, which is reflected in the number of GNN layers. \revised{Since the GNN layer count} is a positive integer, we define the index of each action in the action space as the corresponding action value.

$\bullet$ Policy ($\pi$): The policy in timestep $i$ outputs action $a_{i}$ according to the input state $s_{i}$. Here we apply double deep q-network (DDQN) presented in Sec.~\ref{Sec_Problem4} to simulate and train the policy and call the trained policy a meta-policy.

$\bullet$ Reward function ($REW$): The reward function outputs the reward $r_{i}$ in timestep $i$. Since we expect to improve network representation performance through policy-guided aggregations, we intuitively define the current immediate reward $r_{i}$ as the difference (a decimal) between the current validation classification performance and the performance of the previous timestep.

$\bullet$ Return ($\mathcal{R}$): The return $\mathcal{R}_{i}$ in timestep $i$ indicates the discounted accumulation of all rewards in the interval $[i,t]$. Based on the DDQN, we approximate the Q values output by the DNNs in the DDQN to the rewards over different actions. Since DDQN always chooses the action that maximizes return, it aligns with the goal of standard MDP.

According to these definitions, the process of the meta-policy module in each timestep $i$ includes five stages: 1) Sample a brain network and take its adjacency matrix as the current state $s_{i}$. 2) Determine the number of layers of the GNN that processes the current brain network according to the action $a_{i}$ corresponding to the maximum Q value output by the policy $\pi$. 3) Calculate the current reward $r_{i}$ based on performance changes (technical details will be introduced in the following subsection). 4) Obtain the next state $s_{i+1}$ with a new heuristic strategy for state transition. Concretely, we abstract the brain network of each subject as a coarse node and construct a subject network according to the network building module, where the initial node features are obtained by vectorizing the weighted matrices. Then we realize state transition through node sampling. For example, given the current state $s_{i}$ and action $a_{i}$, we randomly sample a $a_{i}$-hop neighbor of the coarse node corresponding to state $s_{i}$ in the subject's network, \revised{where brain instance's adjacency matrix} corresponding to the sampled neighbor is the next state $s_{i+1}$. In this way, the state transition obeys Markov, i.e., the next state $s_{i+1}$ is only affected by the \revised{current one $s_{i}$ without considering previous states.} 5) Record the process of this timestep and train the policy $\pi$ according to Eq.~\ref{eq_Policy_Loss} and the back-propagation algorithm.

\begin{algorithm}[t]
\small
{\bf Input:}
Brain network dataset $D$, number of timesteps $t$, number of all possible actions $b$, discount coefficient $\gamma$, epsilon probability $\epsilon$, window size of the history records $w$.
\begin{algorithmic}[1]
\State Generate adjacency matrices of brain networks and the subject network via Eq.~(\ref{eq_NetBuild}), $\mathbf{A}(i,j) = exp(-\left\|\mathbf{F}^{(0)}(i)-\mathbf{F}^{(0)}(j)\right\|)$ or 0. \\
Initialize two DNNs in DDQN and two GNNs with $b$ layers.
\State Randomly sample a brain network from $D_{train}$ and get the starting state according to the adjacency matrix.
\For{$i = 1,2,\dots,t$}
    \State Select the action via Eq.~(\ref{eq_Action}), $a_{i} = argmax_{a_{i}}(Q_{eval}(s_{i},a_{i}))$ or a random action.
    \State Train the action value guided GNN1 in MDP via Eq.~(\ref{eq_GNN_Loss}), $\mathcal{L}_{GNN}=-\sum_{G_{i}\in D_{train}} log(\mathbf{E}(i)\mathbf{T}^{(l+1)})\mathbf{Y}(i)^{T}$.
    \State Calculate the reward via Eq.~(\ref{eq_Reward}), $r_{i} = REW(s_{i},a_{i})=PER(s_{i},a_{i})-\frac{1}{w}\sum_{i-w}^{i-1}PER(s_{i},a_{i})$.
    \State Sample the next state and record this process.
    \State Train the policy in MDP via Eq.~(\ref{eq_Policy_Loss}), $\mathcal{L}_{policy}=(max(\mathcal{Q}_{target}(s_{i},a_{i}))-max(\mathcal{Q}_{eval}(s_{i},a_{i})))^{2}$.
\EndFor
\State Use the trained policy, meta-policy as meta-knowledge to guide the training and testing of GNN2.
\State Obtain network representations of the test set $D_{test}$ and perform brain analysis tasks.
\end{algorithmic}
\caption{\textbf{BN-GNN:} GNN-based brain network representation learning framework}
\label{alg_BN-GNN}
\end{algorithm}

\subsection{GNN Module}
\label{Sec_GNN_Module}
The GNN module contains two GNNs with a pooling layer to learn brain network representations. The first GNN (called GNN1) is used in the MDP to train the policy $\pi$. As defined in Sec.~\ref{Sec_Policy_Module}, each action is a positive integer in the interval $[1,b]$, where $b$ is the total \revised{count} of all possible actions. Since the action $a_{i}$ specifies the number of feature aggregations and GNN achieves different aggregations by controlling the number of layers, GNN1 needs to stack $j$ neural networks when $a_{i}=j$ ($j$ is the index of $a_{i}$ in $A$). Considering that the actions in different processes are usually different, reconstructing GNN1 in each timestep is very time- and space-consuming. To alleviate this problem, we use a parameter sharing mechanism to construct a $b$-layer GNN1. For example, given the current action $a_{i}=j$, we only use the first $j$ layers of GNN1 to learn the current brain network $G_{i}$. The aggregation process realized by \revised{GCN}~\cite{kipf2016semi} is as follows:
\begin{equation}
    \small
	\begin{split}
	\mathbf{F}^{(j)}=ReLU(\widehat{\mathbf{A}}\cdots ReLU(\widehat{\mathbf{A}}\mathbf{F}^{(0)}\mathbf{T}^{(1)})\cdots \mathbf{T}^{(j)})
	\end{split}.
	\label{eq_BN-GNN}
\end{equation}
After obtaining the final node feature matrix $\mathbf{F}^{(j)}$, we apply the pooling of Eq.~(\ref{eq_Pool}) to obtain the network representation. Then we use the back-propagation algorithm of Eq.~(\ref{eq_GNN_Loss}) to train GNN1. Since the current timestep only involves the first $b$ layers of GNN1, only the parameters of the first $b$ layers are updated. Compared with constructing a GNN for each network separately in each timestep, the parameter sharing mechanism significantly improves the training efficiency.

To calculate the current reward $r_{i}$, we measure the classification performance of GNN1 on the validation set $D_{val}$. The immediate reward \revised{in MDP is obtained} as follows: 
\begin{equation}
    \small
	\begin{split}
	r_{i} = REW(s_{i},a_{i})=PER(s_{i},a_{i})-\frac{1}{w}\sum_{i-w}^{i-1}PER(s_{i},a_{i}),
	\end{split}
	\label{eq_Reward}
\end{equation}
where $PER$ represents the performance metric of the classification result on the validation data (here we apply accuracy). $w$ indicates the number of historical records used to determine benchmark performance $\frac{1}{w}\sum_{i-w}^{i-1}PER(s_{i},a_{i})$. Compared with only considering the performance of the previous timestep $(i-1)$, the benchmark based on multiple historical performances improves the reliability of $r_{i}$.

Since the training of GNN1 and policy in MDP are usually not completed in the same timestep, it is inconvenient and inappropriate to use GNN1 to perform brain analysis tasks on the test set $D_{test}$. Therefore, after MDP, we apply the trained meta-policy to guide the training and testing of a new GNN (called GNN2), where GNN2 and GNN1 have the same aggregation type and parameter sharing mechanism. The detailed \revised{steps} of BN-GNN is presented in Algorithm~\ref{alg_BN-GNN}.

\section{Experiments} 
\label{Sec_Experiments}
\revised{Eight real-world brain network datasets are used to evaluate our proposed BN-GNN. Our first step is to give the brain analysis dataset information} (Sec.~\ref{Sec_Dataset}), the comparison baselines, and the experimental settings (Sec.~\ref{Sec_Baseline}). We then conduct sufficient experiments on the brain network classification task to \revised{address multiple research questions (RQs) about BN-GNN's effectiveness:}

$\bullet$ RQ1. \revised{Is BN-GNN better than other advanced brain network representation algorithms?} (Sec.~\ref{Sec_Comparison})

$\bullet$ RQ2. Can the three modules included in BN-GNN improve brain network representations learning? (Sec.~\ref{Sec_Ablation})

$\bullet$ RQ3. How do important hyperparameters in BN-GNN affect model representation performance? (Sec.~\ref{Sec_Hyperparameter})

\begin{table}[t]
\caption{Statistics of Brain Network Datasets.}
\vspace{-1mm}
\small
\centering
\begin{tabular}{ccccccccc}
\hline
Dataset           & BP-DTI & BP-fMRI & HIV-DTI & HIV-fMRI & ADHD-fMRI    & HI-fMRI      & GD-fMRI      & HA-EEG   \\ \hline
Network Instances & 70     & 70      & 97      & 97       & 83      & 79      & 85      & 61    \\
Healthy/Male/Active      & 35     & 35      & 45      & 45       & 46      & 44      & 36      & 21    \\
Patient/Female/Passive    & 35     & 35      & 52      & 52       & 37      & 35      & 49      & 40    \\
Nodes$\times$Nodes      & 82$\times$82   & 82$\times$82 & 90$\times$90  & 90$\times$90    & 200$\times$200 & 200$\times$200 & 200$\times$200 & 68$\times$68 \\ \hline
\end{tabular}
\label{tab_Dataset}
\vspace{-1mm}
\end{table}

\subsection{Datasets}
\label{Sec_Dataset}
\revised{The brain analysis tasks consist of eight brain network datasets covering different neurological disorders.}  Table~\ref{tab_Dataset} shows \revised{the internal information of all the data. More specifically, they are collected and processed as detailed below:}

\textbf{Human Immunodeficiency Virus Infection (HIV-DTI \& HIV-fMRI):} 
\revised{~\cite{ragin2012structural} collected this raw data from two modalities, i.e., DTI and fMRI. The subject data contains 70 instances, where half the patients and half are healthy, with similar profiles in age, gender, education level, etc. Following~\cite{ma2017multi}, we employ DPARSF~\cite{yan2010dparsf} to preprocess the fMRI data and then perform Gaussian smoothing on the images. To eliminate noise and drift, we also apply linear trending and bandpass filtering techniques. Further, we divide any instance into 116 ROIs by AAL atlas~\cite{tzourio2002automated} and discard 26 of them. We then harvest the network initial weighting matrices for all subjects' brain instances. For DTI, we first utilize the FSL~\cite{smith2004advances} with techniques such as noise filtering, image correction, etc. to process DTI data.} Then we obtain the corresponding weighted matrices of brain networks with 90 regions. 

\textbf{Bipolar Disorder (BP-DTI \& BP-fMRI):} \revised{The dataset, which also contains fMRI and DTI modalities, includes 45 healthy subjects and 52 bipolar patients with similar profiles}~\cite{cao2015identification}. For \revised{fMRI, we employ the CONN}~\cite{whitfield2012conn} to get the initial brain networks. Concretely, We first realign and co-register the original EPI images and then perform normalization and smoothing. \revised{After that, confounding influences caused by motion artifacts, cerebrospinal fluid, etc. will disappear. In the end, each initial connectome is calculated from the marked gray matter area.} For the DTI data, we follow the data processing strategy in~\cite{ma2017multi} to generate brain networks whose regions are the same as those of the fMRI network. 

\textbf{Attention Deficit Hyperactivity Disorder (ADHD-fMRI) \& Hyperactive Impulsive Disorder (HI-fMRI) \& Gender (GD-fMRI):} 
The initial data was constructed from the whole brain fMRI atlas~\cite{craddock2012whole}. Following the work in~\cite{pan2016task}, we use the functional segmentation result CC200 from~\cite{craddock2012whole}, which divides each \revised{instance into two hundred ROIs}. In order to explore the relationship between ROIs, we record the average value of each ROI in a specific voxel time course. Similarly, we obtain the correlation between the two ROIs according to the Pearson correlation between the two time courses, and generate three reliable brain network instance sets \revised{with the threshold specified in}~\cite{pan2016task}. 
\revised{More processing is stated in}~\cite{craddock2012whole,pan2016task}.

\textbf{Hearing Activity (HA-EEG):} 
The raw electroencephalogram (EEG) data was recorded from 61 healthy adults using 62 electrodes~\cite{hernandez2021perceptual}. The participants were either actively listening to individual words over headphones (active condition) or watching a silent video and ignoring the speech (passive condition). To transform the dataset into a usable version, we perform source analysis using the fieldtrip toolkit~\cite{oostenveld2011fieldtrip} with a cortical-sheet based source model and a boundary element head model. Specifically, we calculate the coherence of all sources and segment the sources based on the 68 regions of the Desikan-Killiany cortical atlas. Furthermore, we utilize the imaginary part of the coherence spectrum as the connectivity metric to reduce the effect of electric field spread~\cite{nolte2004identifying}.

\begin{figure*}[t]
	\raggedleft
	\subfloat[Weighted matrices $\mathbf{W}$]{
    \begin{minipage}[t]{0.19\linewidth}
    \centering
    \includegraphics[width=1.\linewidth]{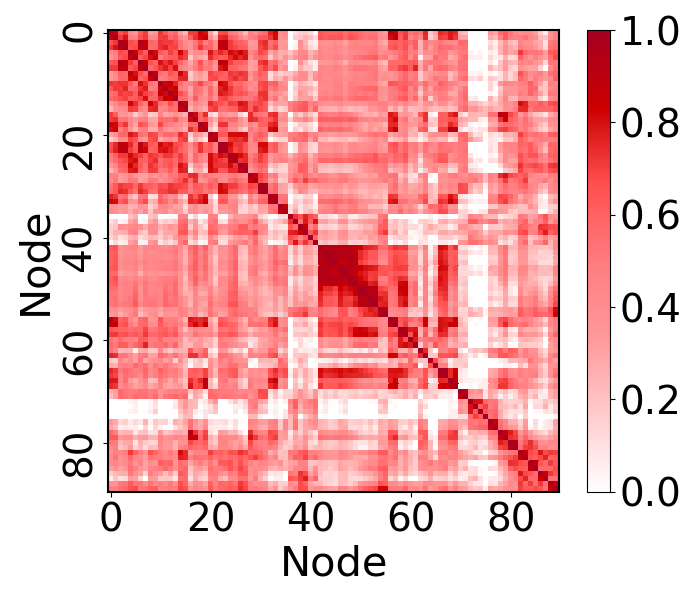}\\
    \includegraphics[width=1.\linewidth]{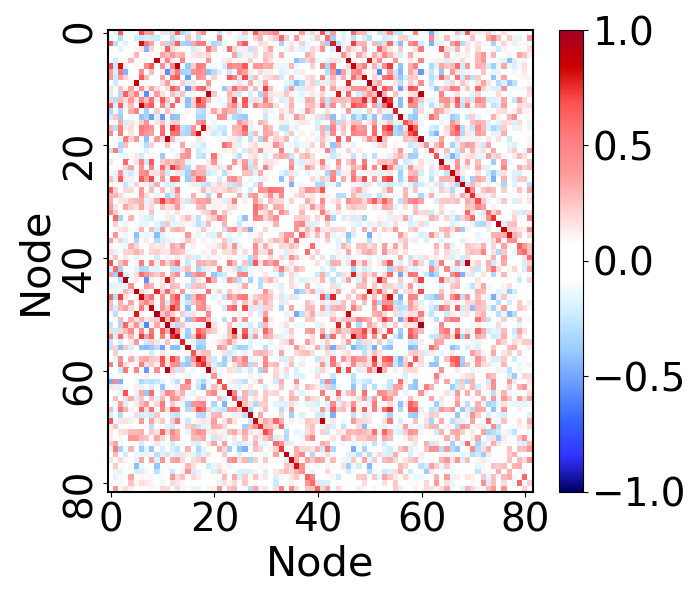}
    \end{minipage}
    }
	\subfloat[Adjacency matrices $\mathbf{A}$]{
    \begin{minipage}[t]{0.19\linewidth}
    \centering
    \includegraphics[width=1.\linewidth]{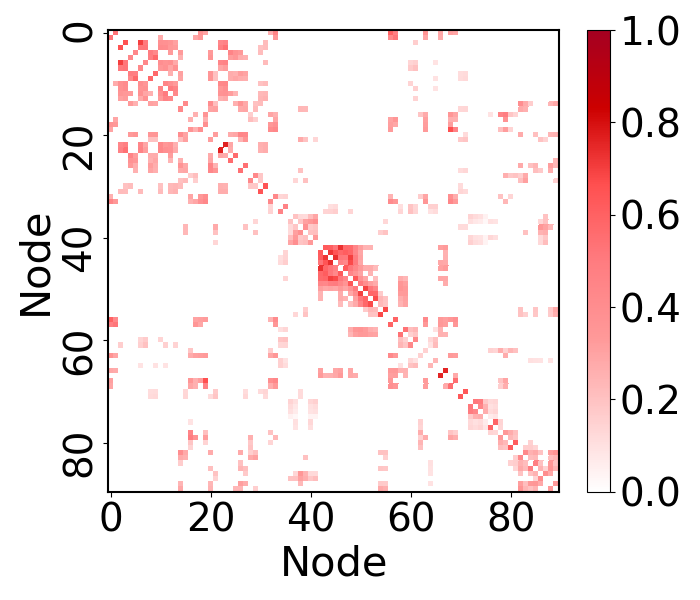}\\
    \includegraphics[width=1.\linewidth]{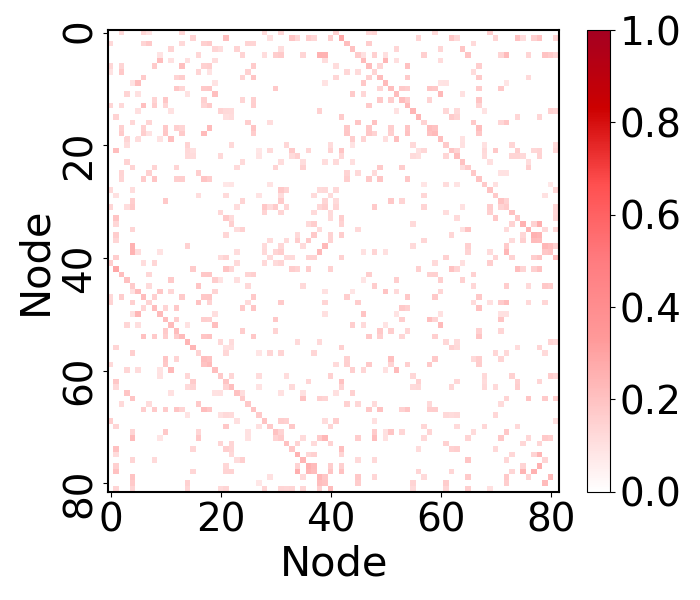}
    \end{minipage}
    }
	\subfloat[Matrices with self-loop $\widetilde{\mathbf{A}}$]{
    \begin{minipage}[t]{0.19\linewidth}
    \centering
    \includegraphics[width=1.\linewidth]{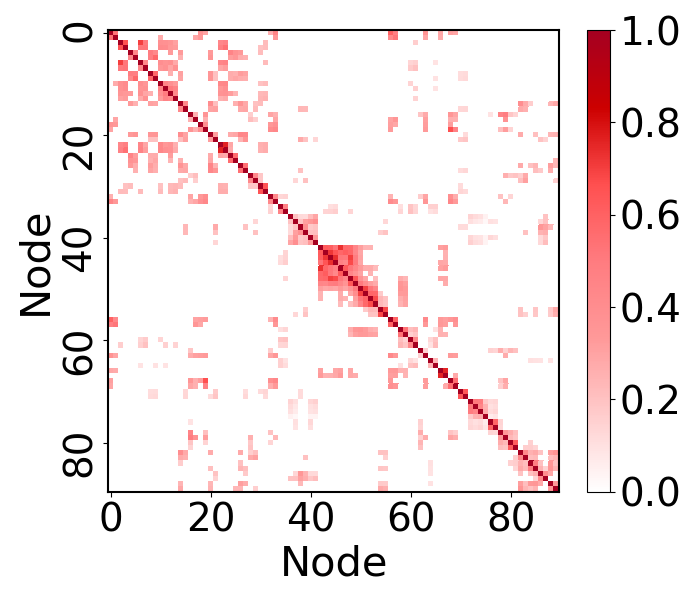}\\
    \includegraphics[width=1.\linewidth]{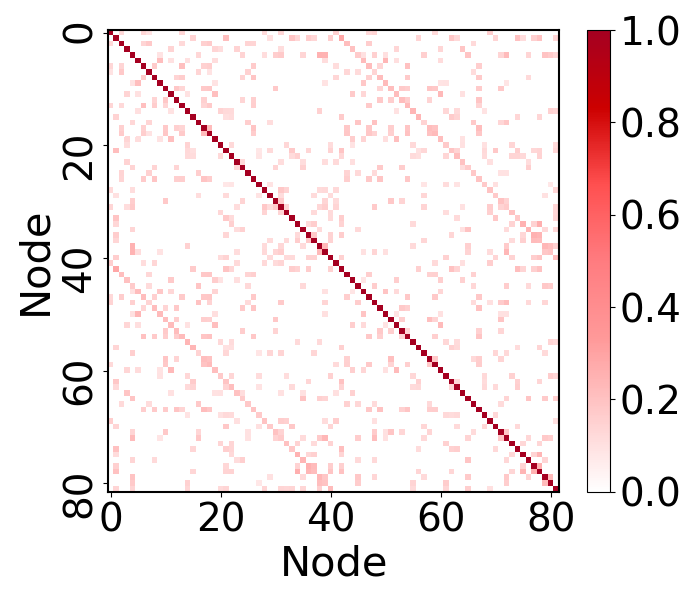}
    \end{minipage}
    }
	\subfloat[Degree matrices $\mathbf{D}$]{
    \begin{minipage}[t]{0.19\linewidth}
    \centering
    \includegraphics[width=1.\linewidth]{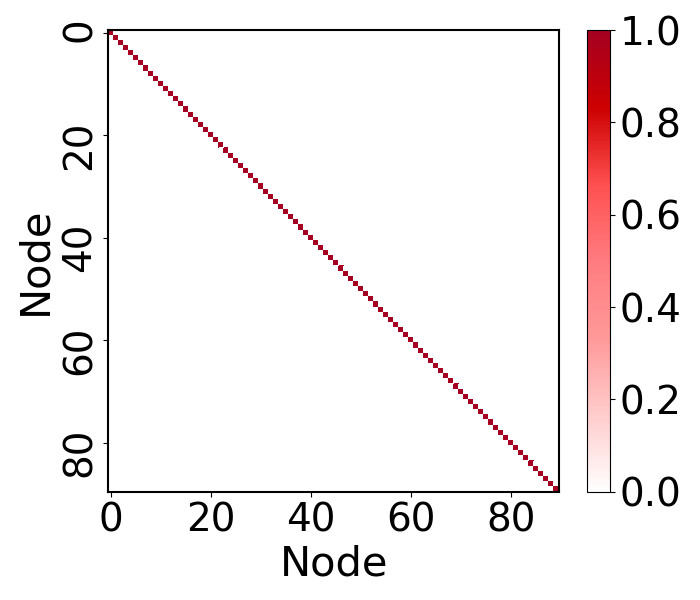}\\
    \includegraphics[width=1.\linewidth]{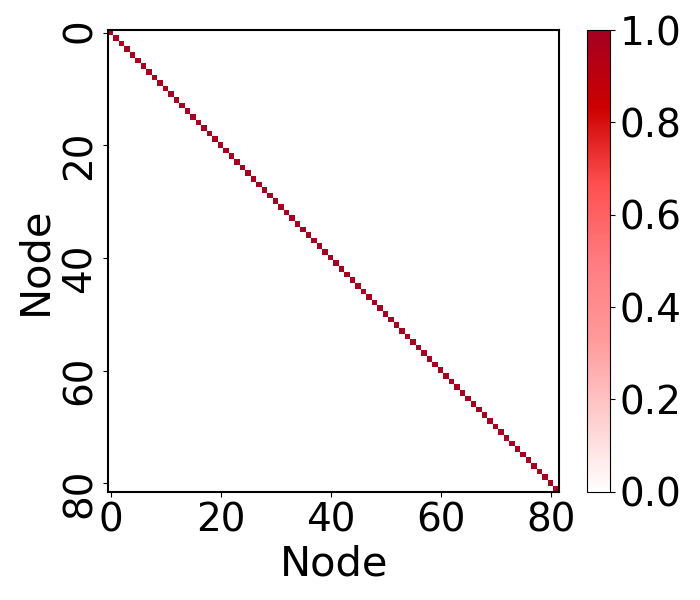}
    \end{minipage}
    }
	\subfloat[Normalized matrices $\widehat{\mathbf{A}}$]{
    \begin{minipage}[t]{0.19\linewidth}
    \centering
    \includegraphics[width=1.\linewidth]{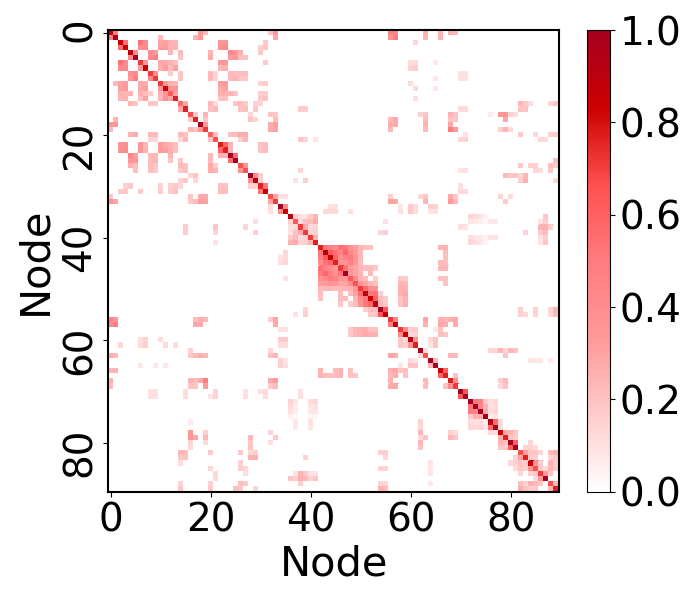}\\
    \includegraphics[width=1.\linewidth]{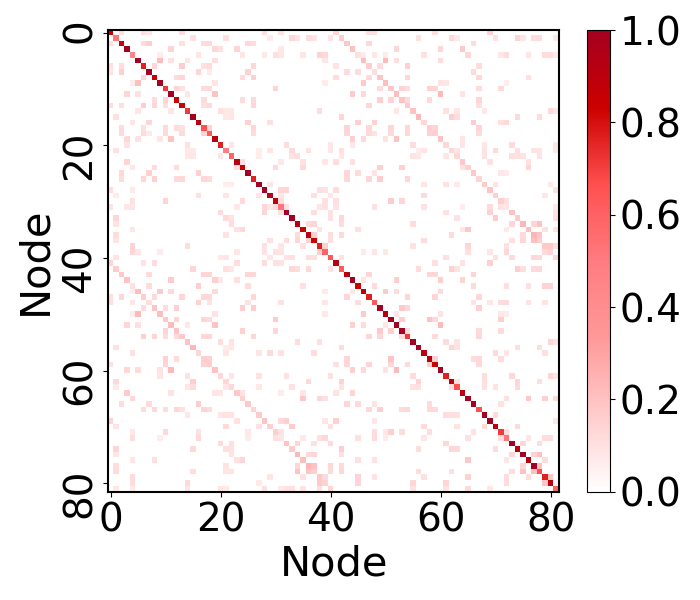}
    \end{minipage}
    }
    \vspace{-2mm}
	\caption{Examples of brain networks of two subjects in fMRI modality. The subject above is from HIV, while below one is from BP. From left to right are the initial weighted matrices $\mathbf{W}$, the adjacency matrices $\mathbf{A}$ generated by the network building module (Eq.~\ref{eq_NetBuild}), and the other matrices involved in the GCN aggregation process (Eq.~\ref{eq_GCN}).}
	\label{fig_Matrix}
	\vspace{-1mm}
\end{figure*}

\subsection{Baselines and Settings}
\label{Sec_Baseline}
\revised{To evaluate BN-GNN, we compare it with multiple excellent baselines whose information is as follows:}

\textbf{DeepWalk \& Node2Vec}~\cite{perozzi2014deepwalk,grover2016node2vec}: The main idea of Deepwalk is to perform random walks in the network, then generate a large number of node sequences, further input these node sequences as samples into word2vec~\cite{mikolov2013efficient}, and finally obtain \revised{meaningful node representation vectors.} Compared with DeepWalk, Node2Vec balances the homophily and structural equivalence of the network through biased random walks. Both of them are commonly used baselines in network representation learning. 

\textbf{GCN \& GAT}~\cite{kipf2016semi,velivckovic2017graph}: Graph convolutional network (GCN) performs convolution aggregations in the graph Fourier domain, while graph attention network (GAT) performs aggregations in combination with the attention mechanism. Both of them are \revised{outstanding GNNs}. 

\textbf{GCN+skip \& GAT+skip}: \revised{Following}~\cite{li2019deepgcns}, we construct GCN+skip and GAT+skip by adding residual skip-connections to GCN and GAT, respectively. 

\textbf{GraphSAGE \& FastGCN}~\cite{Hamilton2017inductive,chen2018fastgcn}: They are two improved GNN algorithms with different sampling strategies. For the sake of computational efficiency, GraphSAGE only samples \revised{a predefined number of neighbor nodes as objects to aggregate. Unlike GraphSAGE, which samples neighbor nodes, FastGCN samples all nodes, constructs a new topology based on the initial structure and encodes global information.}

\textbf{PR-GNN \& GNEA \& Hi-GCN}~\cite{li2020pooling,bi2020gnea,jiang2020hi}: Three GNN-based baselines for brain network analysis, all of which contain methods for optimizing the initial brain network generated by neuroimaging technology. PR-GNN utilizes the regularized pooling layers to filter nodes in the network and uses GAT for feature aggregation. GNEA \revised{determines a constant number of neighbors for all nodes through} the correlation coefficient in each brain network. Hi-GCN uses the eigenvector-based pooling layers EigenPooling to generate multiple coarse-grained sub-graphs from the initial network and then aggregates network information hierarchically and generates network representations. 

\textbf{SDBN}~\cite{wang2017structural}: \revised{Instead of involving GNNs, it introduces convolutional neural networks (CNNs)~\cite{krizhevsky2012imagenet} to perform connectome embedding for subjects' brain instances.}

\revised{For settings, we complete BN-GNN with} GCN and GAT, respectively, namely \textbf{BN-GCN and BN-GAT}.
Moreover, we set the total number of timesteps $t$ to 1000, the total number of all possible actions $b$ to 3, the window size $w$ of $REW$ to 20, and the discount coefficient $\gamma$ to 0.95. For the epsilon probability $\epsilon$, we set it to decrease linearly in the first 20 timesteps, with a starting probability of 1.0 and an ending probability of 0.05. For all GNN-based methods, we use $ReLU$ with a slope of 0.2 as the activation \revised{operator of feature aggregations and use dropout with a rate} of 0.3 between every two adjacency neural networks. For a fair comparison, we use $Adam$ optimizers with learning rates of 0.0005 and 0.005 to \revised{update the policy and GNN2.} We set the network representation dimension of all methods to 128 and employ the strategies mentioned in the corresponding papers to adjust the parameters of \revised{baselines and show results} with the best settings. Besides, we use the same data split ($|D_{trian}|:|D_{val}|:|D_{test}|=8:1:1$) to repeat each experiment 10 times, where each experiment records the test result with the highest verification value within 100 epochs. All experiments are performed on the same server with two 20-core CPUs (126G) and an NVIDIA Tesla P100 GPU (16G).

\renewcommand{\arraystretch}{1.05}
\begin{table}[t]
\caption{\revised{Performance comparison of multiple algorithms} on brain network classification tasks. 
The first part compares the experimental results of multiple methods, while the second and third parts refine the representation performance of GCNs and GATs with different layers.
The bold and italicized values in each part represent the best and second-best results of all methods.
\textbf{$\uparrow$} indicates the improvement (\%) of our BN-GNN compared to the best baseline of each part.}
\vspace{-1mm}
\small
\centering
\resizebox{\linewidth}{!}{
\begin{tabular}{cccccccccc}
\hline
Method    & Layers   & BP-DTI               & BP-fMRI              & HIV-DTI              & HIV-fMRI             & ADHD-fMRI            & HI-fMRI              & GD-fMRI              & HA-EEG               \\ \hline
DeepWalk  & -        & 0.520±0.097          & 0.530±0.134          & 0.514±0.159          & 0.485±0.130          & 0.512±0.141          & 0.462±0.148          & 0.550±0.127          & 0.566±0.270          \\
Node2Vec  & -        & 0.530±0.110          & 0.550±0.111          & 0.514±0.145          & 0.500±0.172          & 0.525±0.075          & 0.475±0.122          & 0.562±0.170          & 0.583±0.200          \\
PR-GNN    & 2        & 0.590±0.186          & \textit{0.630±0.110} & 0.557±0.174          & 0.585±0.100          & \textit{0.625±0.167} & \textit{0.600±0.165} & 0.600±0.145          & 0.650±0.157          \\
GNEA      & 3        & 0.560±0.149          & 0.600±0.134          & 0.557±0.118          & 0.585±0.196          & 0.550±0.127          & 0.562±0.160          & \textit{0.612±0.087} & 0.633±0.221          \\
HI-GCN    & 3        & 0.540±0.162          & 0.600±0.109          & 0.528±0.192          & 0.571±0.127          & 0.562±0.128          & 0.562±0.160          & 0.587±0.148          & 0.616±0.183          \\
GraphSAGE & 2        & \textit{0.610±0.192} & 0.610±0.113          & 0.571±0.202          & 0.600±0.124          & 0.575±0.127          & 0.575±0.100          & 0.600±0.165          & \textit{0.716±0.076} \\
FastGCN   & 2        & 0.590±0.113          & 0.620±0.140          & \textit{0.585±0.134} & \textit{0.628±0.145} & 0.612±0.180          & 0.600±0.145          & 0.600±0.175          & 0.700±0.194          \\
BN-GNN    & 1$\sim$3 & \textbf{0.630±0.167} & \textbf{0.640±0.120} & \textbf{0.614±0.111} & \textbf{0.642±0.146} & \textbf{0.637±0.130} & \textbf{0.612±0.205} & \textbf{0.637±0.141} & \textbf{0.733±0.200} \\
Gain    & -        & \textbf{2.0$\,\uparrow$}       & \textbf{1.0$\,\uparrow$}       & \textbf{2.9$\,\uparrow$}       & \textbf{1.4$\,\uparrow$}       & \textbf{1.2$\,\uparrow$}       & \textbf{1.2$\,\uparrow$}       & \textbf{2.5$\,\uparrow$}       & \textbf{1.7$\,\uparrow$}       \\ \hline
GCN       & 1        & 0.560±0.101          & 0.600±0.148          & 0.542±0.178          & 0.557±0.174          & 0.562±0.150          & \textit{0.587±0.080} & \textit{0.600±0.122} & 0.650±0.189          \\
GCN       & 2        & \textit{0.590±0.130} & 0.610±0.122          & 0.571±0.180          & \textit{0.600±0.166} & \textit{0.600±0.165} & 0.587±0.125          & 0.600±0.108          & 0.666±0.129          \\
GCN       & 3        & 0.540±0.180          & 0.600±0.184          & 0.542±0.189          & 0.585±0.118          & 0.525±0.122          & 0.562±0.187          & 0.575±0.150          & 0.616±0.106          \\
GCN+skip  & 3        & 0.590±0.164          & \textit{0.620±0.116} & \textit{0.585±0.134} & 0.528±0.111          & 0.562±0.170          & 0.575±0.160          & 0.600±0.183          & \textit{0.700±0.163} \\
BN-GCN    & 1$\sim$3 & \textbf{0.610±0.170} & \textbf{0.640±0.120} & \textbf{0.614±0.111} & \textbf{0.642±0.172} & \textbf{0.637±0.130} & \textbf{0.600±0.165} & \textbf{0.625±0.125} & \textbf{0.716±0.197} \\
Gain    & -        & \textbf{2.0$\,\uparrow$}       & \textbf{2.0$\,\uparrow$}       & \textbf{2.9$\,\uparrow$}       & \textbf{4.2$\,\uparrow$}       & \textbf{3.7$\,\uparrow$}       & \textbf{1.3$\,\uparrow$}       & \textbf{2.5$\,\uparrow$}       & \textbf{1.6$\,\uparrow$}       \\ \hline
GAT       & 1        & 0.570±0.141          & \textit{0.620±0.172} & 0.542±0.261          & 0.571±0.169          & 0.562±0.128          & 0.575±0.203          & 0.612±0.189          & 0.650±0.203          \\
GAT       & 2        & 0.590±0.130          & 0.610±0.157          & 0.585±0.149          & \textit{0.614±0.111} & \textit{0.600±0.215} & \textit{0.587±0.137} & 0.612±0.152          & 0.683±0.174          \\
GAT       & 3        & 0.550±0.128          & 0.610±0.144          & 0.571±0.202          & 0.585±0.149          & 0.550±0.127          & 0.575±0.160          & 0.612±0.171          & 0.666±0.235          \\
GAT+skip  & 3        & \textit{0.600±0.184} & 0.610±0.083          & \textit{0.600±0.153} & 0.557±0.162          & 0.575±0.169          & 0.587±0.185          & \textit{0.625±0.111} & \textit{0.700±0.124} \\
BN-GAT    & 1$\sim$3 & \textbf{0.630±0.167} & \textbf{0.640±0.128} & \textbf{0.614±0.181} & \textbf{0.642±0.146} & \textbf{0.612±0.180} & \textbf{0.612±0.205} & \textbf{0.637±0.141} & \textbf{0.733±0.200} \\
Gain    & -        & \textbf{3.0$\,\uparrow$}       & \textbf{2.0$\,\uparrow$}       & \textbf{1.4$\,\uparrow$}       & \textbf{2.8$\,\uparrow$}       & \textbf{1.2$\,\uparrow$}       & \textbf{2.5$\,\uparrow$}       & \textbf{1.2$\,\uparrow$}       & \textbf{3.3$\,\uparrow$}       \\ \hline
\end{tabular}}
\label{tab_Comparison}
\vspace{-1mm}
\end{table}

\subsection{Model Comparison (RQ1)}
\label{Sec_Comparison}
To compare the performance of all methods, we perform disease or gender prediction (i.e., brain network classification) tasks on eight real-world datasets. Moreover, \revised{average accuracy and AUC are utilized as measurement metrics.} Considering that some baselines are challenging to deal with the initial weighted matrices of the brain networks that are almost complete graphs, we perform representation learning for all methods on adjacency matrices generated by the network building module. Taking GCN as an example, Fig.~\ref{fig_Matrix} visualizes the transformation process of the adjacency matrices of two subjects from the HIV-fMRI and BP-fMRI. \revised{According to the values displayed in Table~\ref{tab_Comparison}, five conclusions are reached:}

\begin{figure*}[t]
	\raggedleft
	\subfloat[BP-fMRI]{
    \begin{minipage}[t]{0.49\linewidth}
    \centering
    \includegraphics[width=1.\linewidth]{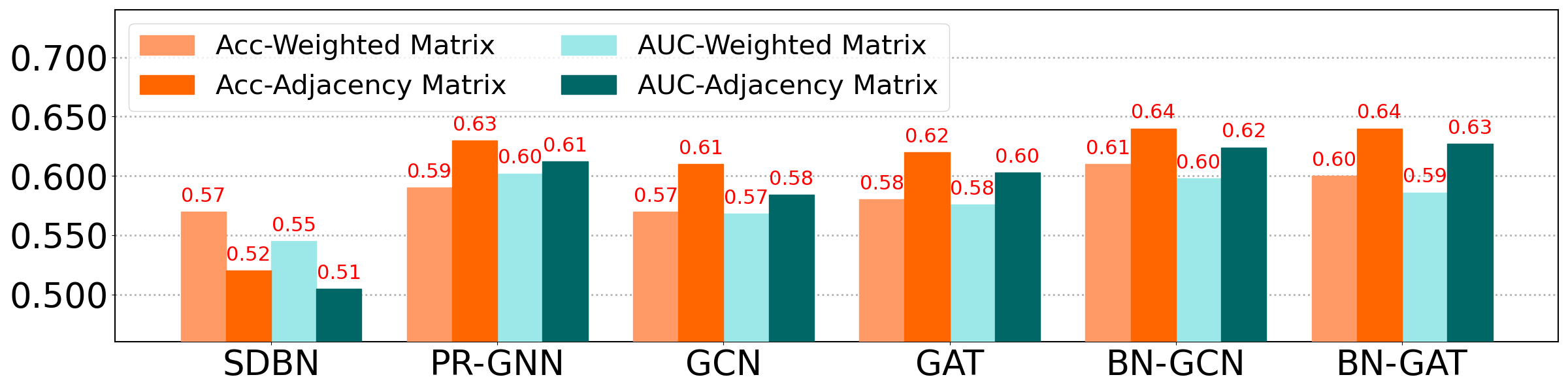}
    \end{minipage}
    }
	\subfloat[HIV-fMRI]{
    \begin{minipage}[t]{0.49\linewidth}
    \centering
    \includegraphics[width=1.\linewidth]{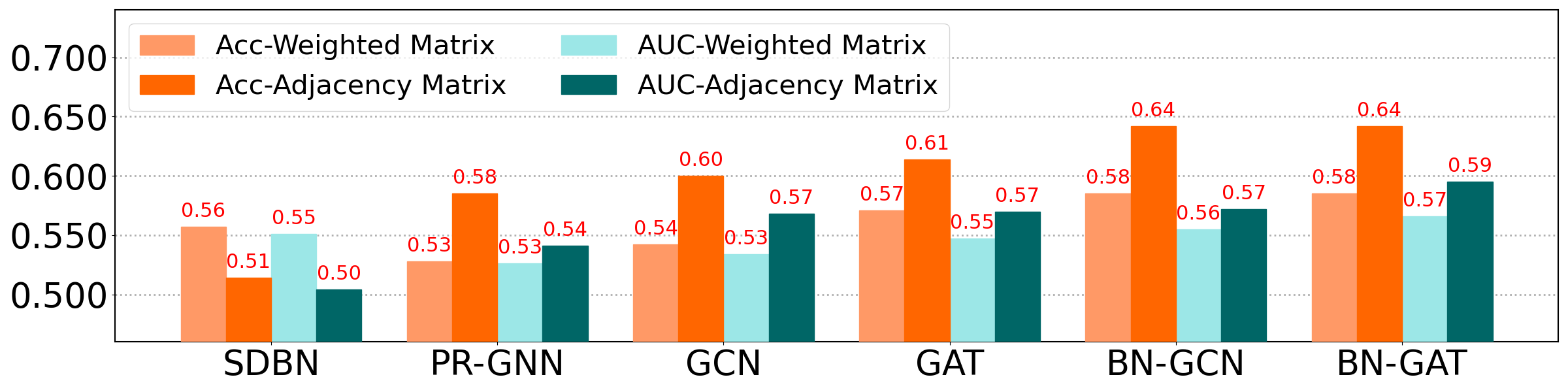}
    \end{minipage}
    }
    \vspace{-2mm}
	\caption{Visualization of ablation experiments to \revised{examine the effectiveness} of the network building module on \revised{BP and HIV}.}
	\label{fig_Bar}
	\vspace{-1mm}
\end{figure*}

\begin{figure*}[t]
	\raggedleft
	\subfloat[ADHD-fMRI]{
    \begin{minipage}[t]{0.24\linewidth}
    \centering
    \includegraphics[width=1.66in]{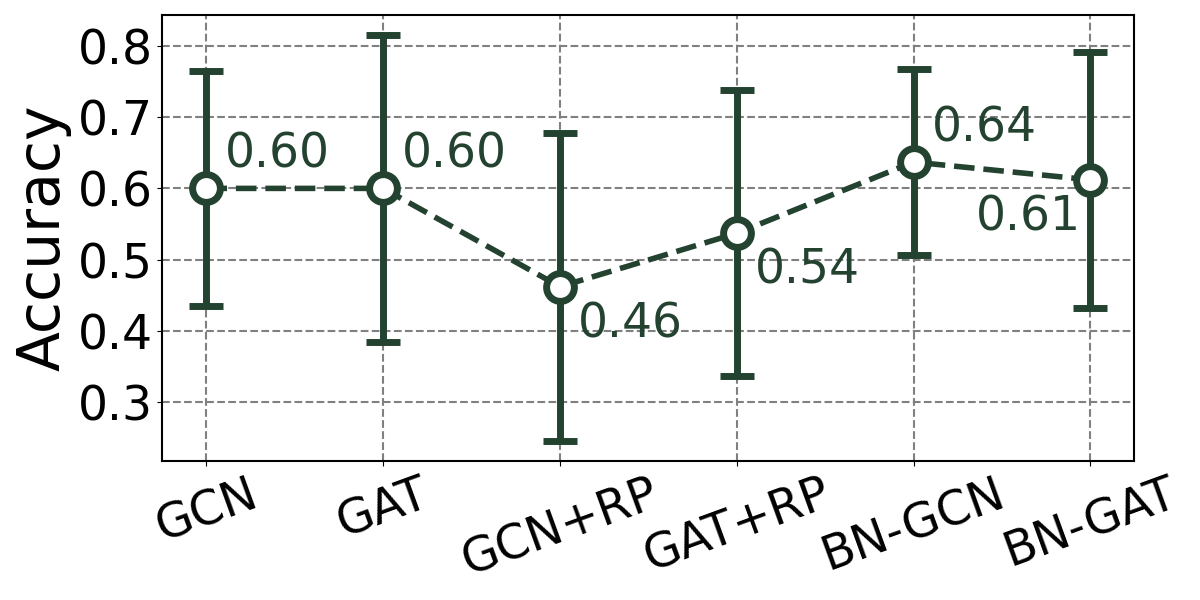}
    \end{minipage}
    }
	\subfloat[HI-fMRI]{
    \begin{minipage}[t]{0.24\linewidth}
    \centering
    \includegraphics[width=1.66in]{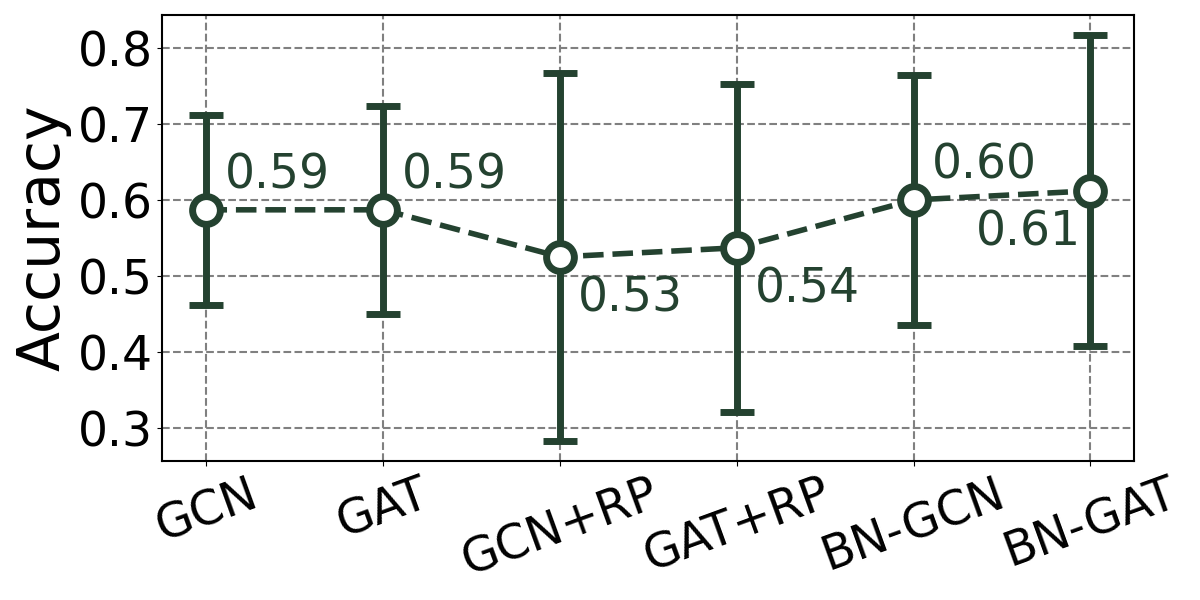}
    \end{minipage}
    }
	\subfloat[GD-fMRI]{
    \begin{minipage}[t]{0.24\linewidth}
    \centering
    \includegraphics[width=1.66in]{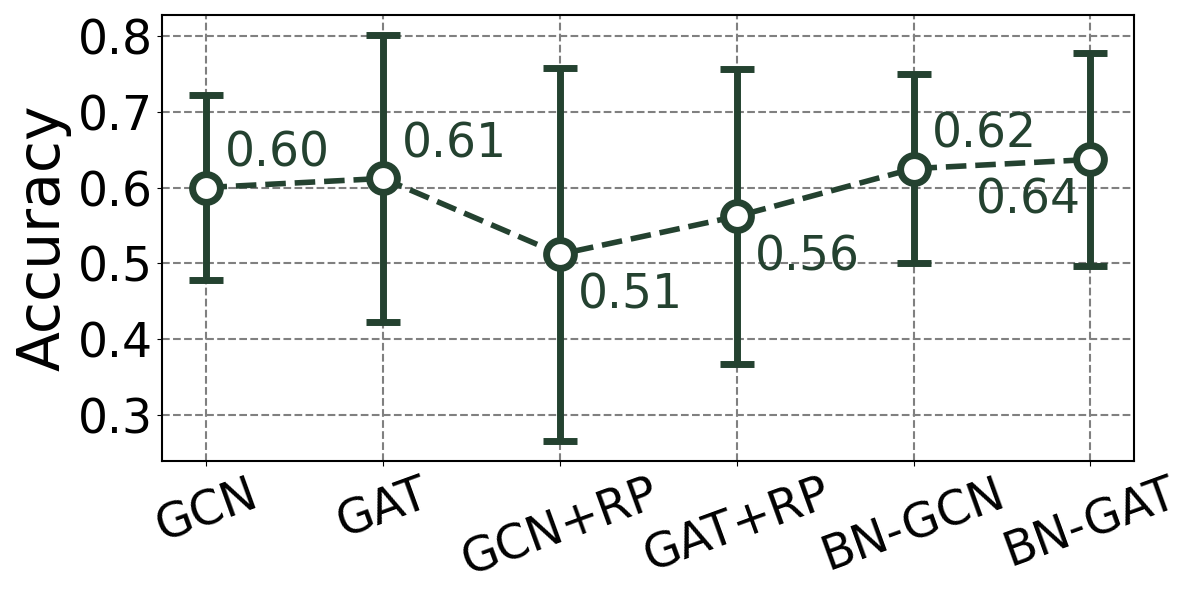}
    \end{minipage}
    }
	\subfloat[HA-EEG]{
    \begin{minipage}[t]{0.24\linewidth}
    \centering
    \includegraphics[width=1.66in]{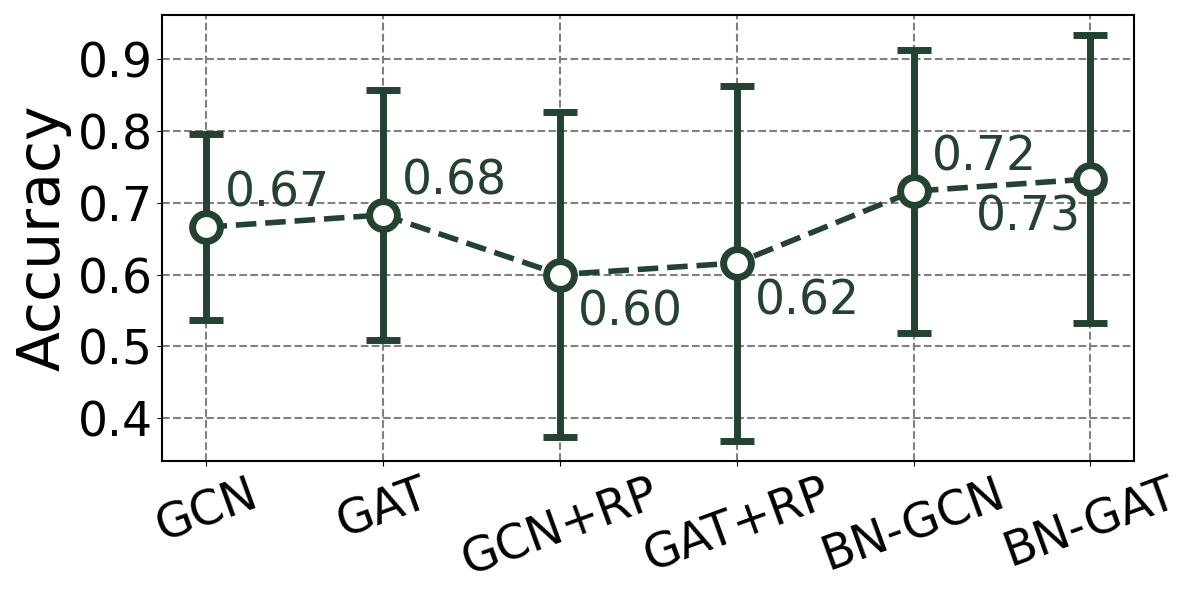}
    \end{minipage}
    }
    \vspace{-2mm}
	\caption{Visualization of ablation experiments to verify the performance of the meta-policy module on four datasets, where GCN+RP and GAT+RP apply a random-policy instead of the meta-policy to guide the aggregation process of GCN and GAT.}
	\label{fig_Errorbar}
	\vspace{-1mm}
\end{figure*}

\begin{figure*}[t]
	\raggedleft
	\subfloat[Input weighted matrices $\mathbf{W}$]{
    \begin{minipage}[t]{0.32\linewidth}
    \centering
    \includegraphics[width=2.2in]{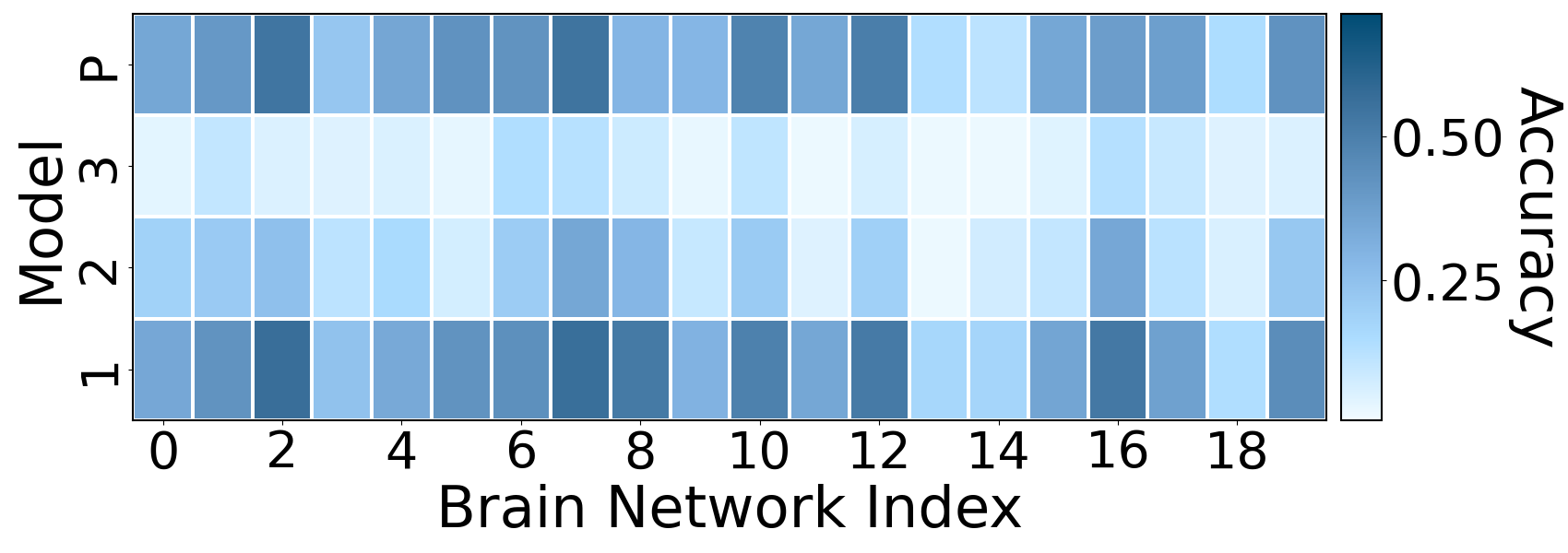}
    \end{minipage}
    }
	\subfloat[Input degree matrices $\mathbf{D}$]{
    \begin{minipage}[t]{0.32\linewidth}
    \centering
    \includegraphics[width=2.2in]{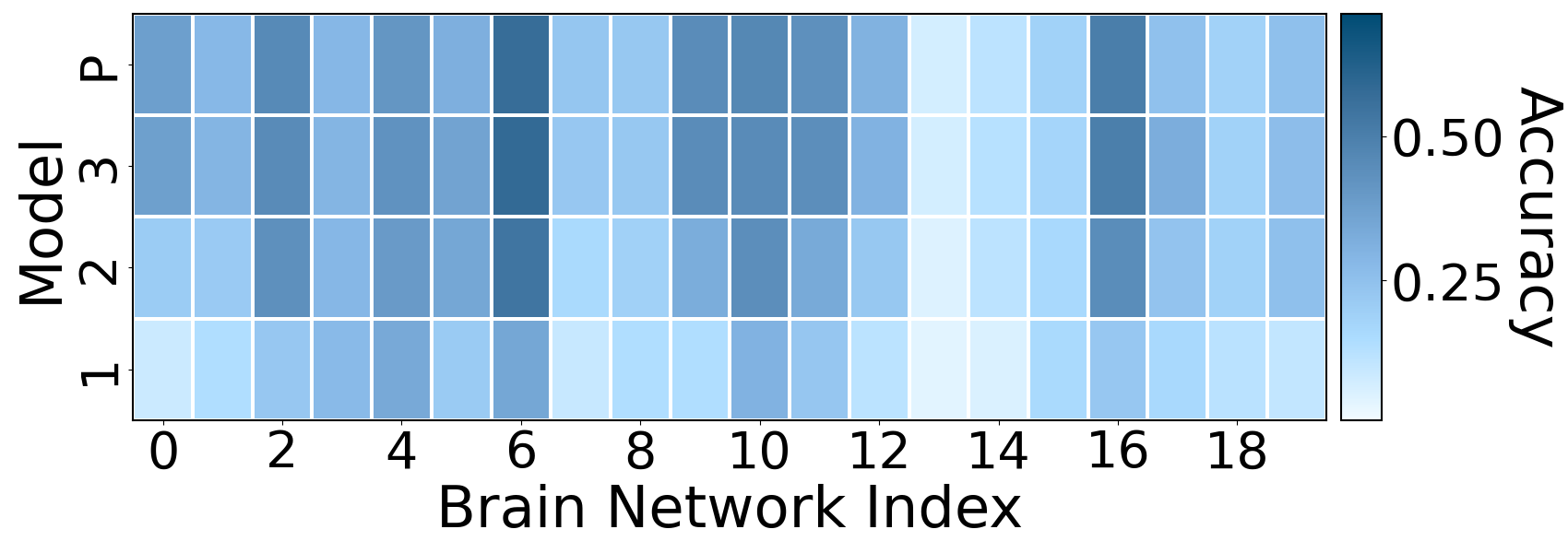}
    \end{minipage}
    }
	\subfloat[Input normalized matrices $\widehat{\mathbf{A}}$]{
    \begin{minipage}[t]{0.32\linewidth}
    \centering
    \includegraphics[width=2.2in]{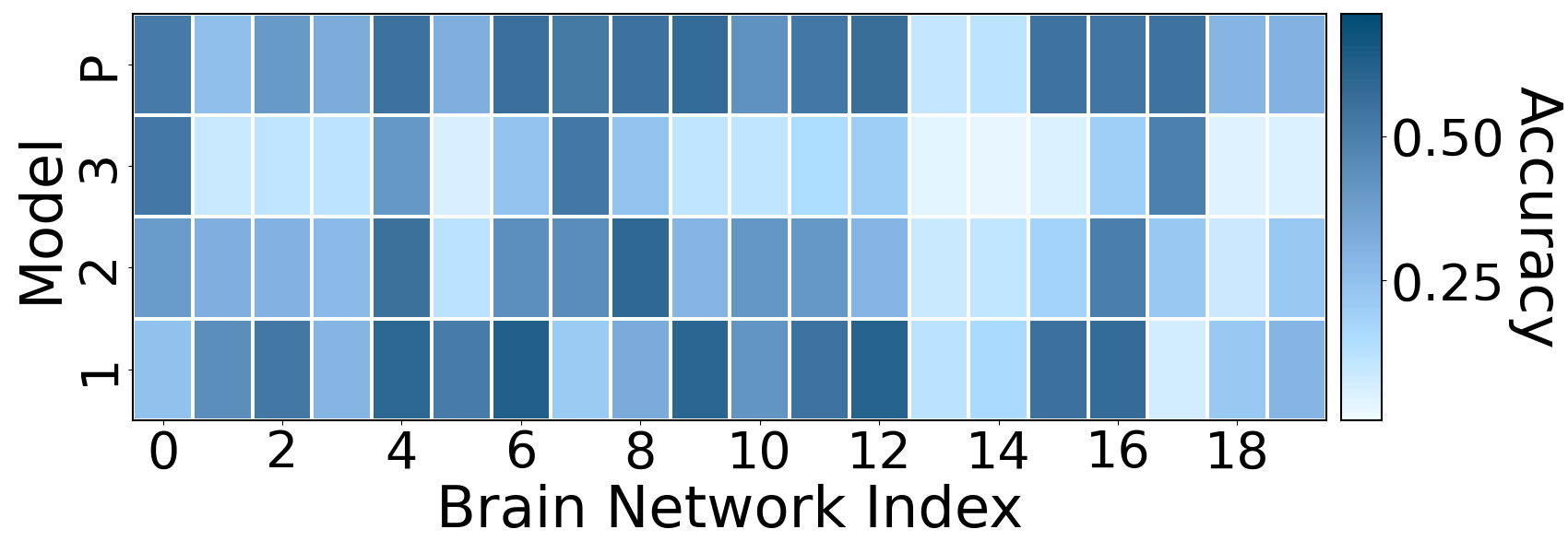}
    \end{minipage}
    }
    \vspace{-2mm}
    \caption{
    Visualization of classification results for 20 brain networks randomly sampled from BP-DTI, where different figures use different types of inputs/adjacency matrices.
    From bottom to top, the $y$-axis represents the GCN model with $y$ layers and the GCN model guided by our meta-policy.
    The $x$-axis represents the brain network index.
    The colors from light to dark mean the average accuracy from low to high.
    }
	\label{fig_Heat}
	\vspace{-1mm}
\end{figure*}

1) BN-GNN always obtains the highest average accuracy value on all datasets, proving that its brain network representation performance is \revised{superior to the baselines.} Specifically, the classification accuracy of BN-GNN on eight datasets is \revised{on average about 2.0\% higher than that of the sub-optimal algorithm.} 2) All GNN-based methods outperform traditional network representation methods (i.e., DeepWalk and Node2Vec). This phenomenon is expected because the GNN architecture can better capture the local \revised{structural features in brain networks, which in turn yield more informative regional representations.} In addition, in brain network classification tasks, end-to-end learning strategies in brain network classification tasks are often superior to unsupervised representation learning methods. 3) GAT-based methods are generally better than GCN-based methods. Compared with the latter, \revised{when layers are stacked over two,} the performance of the former is usually not greatly degraded. This is because the attention mechanism included in GAT alleviates the over-smoothing problem on some datasets. 4) GNNs combined with skip-connections (i.e., GCN+skip and GAT+skip) do not always enable deeper neural networks to perform better. Compared with the best GCN and GAT models \revised{(in the last two parts of Table~\ref{tab_Comparison}), BN-GCN and BN-GAT} have an average accuracy improvement of 2.5\% and 2.1\% on eight classification tasks, respectively. Although these observations reveal the limitations of skip-connections, they also confirm the hypothesis of this work that different brain networks require different aggregation iterations. In other words, since the brain networks of real subjects are usually different, customizing different GNN architectures for different subjects is essential to improve network representation performance and provide therapeutic intervention. 5) Though GraphSAGE and FastGCN improve the efficiency or structure information mining capability of the original GCN, their performance is still inferior to our BN-GNN. This phenomenon indicates that searching suitable feature aggregation strategies for network instances in brain network analysis may be more important than exploring sampling or structural reconstruction strategies. 

\subsection{Ablation Study (RQ2)}
\label{Sec_Ablation}
The classification results and analysis in Sec.~\ref{Sec_Comparison} confirm the superiority of GNN-based methods in processing brain network data.
\revised{Furthermore, we implement ablation studies to detect the independent influence of the network building and meta-policy modules contained in our proposed BN-GNN on the above classification tasks.}
Specifically, for the network building module, we compare the classification results based on the initial weighted matrices and processed adjacency matrices on BP-fMRI and HIV-fMRI, respectively. 
For the meta-policy module, we compare the performance of BN-GNN and random-policy-based GNN on four datasets.
Furthermore, we show examples of practical use of our idea on various input types to see better how it works.

Fig.~\ref{fig_Bar} visualizes the accuracy and AUC scores of ablation studies for network building, from which we \revised{can obtain three key observations:} 1) Utilizing the adjacency matrices generated by the network building module to replace the initial weighted matrices greatly improves the classification performance of the GNN-based methods under the two metrics. This phenomenon indicates that our proposed network building module is beneficial to promote the application of \revised{GNN in brain network analysis studies.} 2) The performance of SDBN on the initial matrices is better than that on the adjacency matrices. On the one hand, SDBN reconstructs the brain networks and \revised{reinforce the spatial structure induction} of the initial weighted matrices, thereby enabling the CNN to capture the highly non-linear features. On the other hand, the sparse adjacency matrices generated by the network building module may not be suitable for CNN-based methods. \revised{Notably, the optimal results of SDBN are} always inferior to that of BN-GNN, indicating that it is meaningful to learn topological brain networks based on GNN. 3) Even though GAT-based methods (including PR-GNN and GAT) use the attention \revised{technique to learn weights for different neighbors}, they are still difficult to deal with densely connected initial brain networks. Thus, generating adjacency matrices (as shown in Fig.~\ref{fig_Matrix}) for GNN \revised{can improve brain network representation learning.}

We replace the meta-policy module in BN-GNN with a random-policy (randomly chooses an action for a given instance) to construct the baselines of ablation experiments, namely GCN+RP and GAT+RP. Fig.~\ref{fig_Errorbar} illustrates the results of ablation experiments for the meta-policy module, from which \revised{two conclusions can be drawn:} 1) The performance of GCN+RP and GAT+RP are worse than BN-GNN and original GNNs on ADHD-fMRI, HI-fMRI, GD-fMRI, and HA-EEG. 2) Our BN-GNN is better than the original GNNs. 
These phenomena once again imply that the introduction of our meta-policy can effectively improve the classification performance of brain networks.

To explore the practical application of our idea on different input types, we illustrate the classification performance of layer-fixed GCNs and GCN guided by meta-policy on BP-DTI in Fig.~\ref{fig_Heat}.
First, we observe Fig.~\ref{fig_Heat}(a) and find that single-layer GCNs generally perform best when using initial weighted matrices as inputs. 
Besides, \revised{GCN performance degrades drastically when its model stack increases.}
This may be because the initial brain networks usually have a high connection density (as shown in Fig.~\ref{fig_Matrix}(a)), which leads to the over-smoothing problem of multi-layer GCNs during the learning process.
Second, Fig.~\ref{fig_Heat}(b) shows the classification results with the degree matrices as inputs.
Since the degree matrices do not encode neighbor information (as shown in Fig.~\ref{fig_Matrix}(b)), these GCNs degenerate into fully-connected neural networks without feature aggregation.
Therefore, the brain network classification performance of models with different numbers of layers varies relatively smoothly.
Finally, after processing the initial data based on our network building module, the optimal number of GCN layers required for normalized matrices with reduced density becomes very different, as shown in Fig.~\ref{fig_Matrix}(c).
In addition, reliable brain networks improve the overall performance upper bound compared to the brain classification of the first two types of inputs.
This again implies the importance of building reliable brain networks and customizing the optimal number of aggregations.
In particular, our meta-policy is often able to find the optimal number of model layers corresponding to brain instances for three input types. 
Therefore, the GCN with our meta-policy (i.e., BN-GCN) generally performs the best (shown in the first row of subfigures in Fig.~\ref{fig_Matrix}), even if the adjacency matrices have extremely large or small connection densities.
In future research, the network building module can continue to serve brain network analysis. 
And the meta-policy may be extended to other scenarios with differentiated requirements for the number of model layers rather than being limited to the application of GNNs.

\begin{figure}[t]
    \centering
    \subfloat[Number of Neighbors]{
    \includegraphics[width=0.32\linewidth]{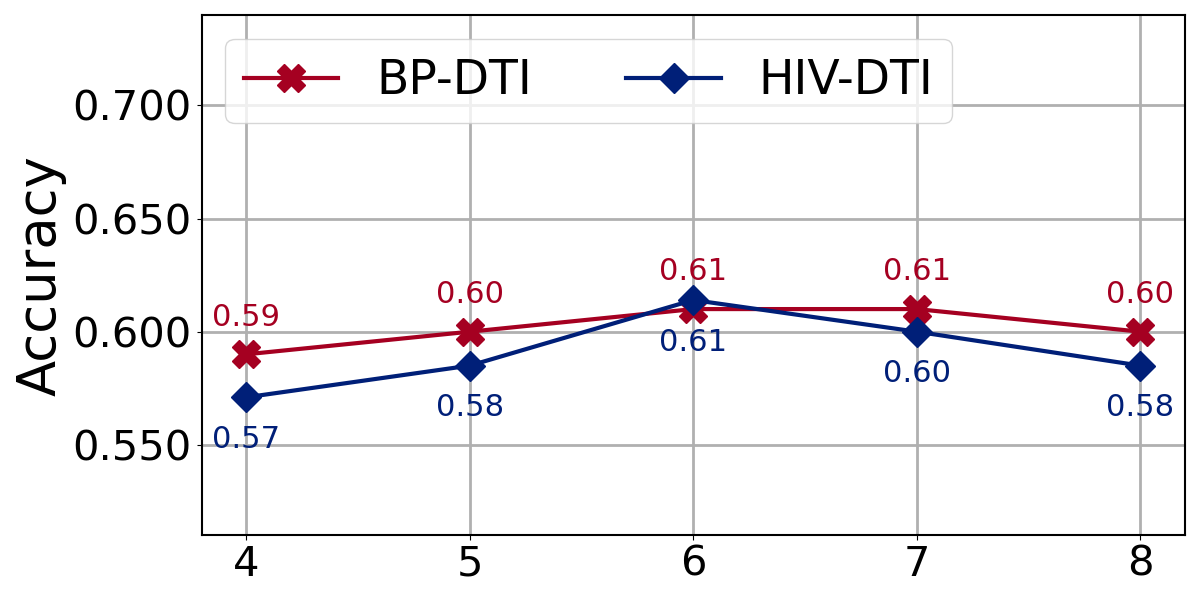}}
    \subfloat[Number of All Possible Actions]{
    \includegraphics[width=0.32\linewidth]{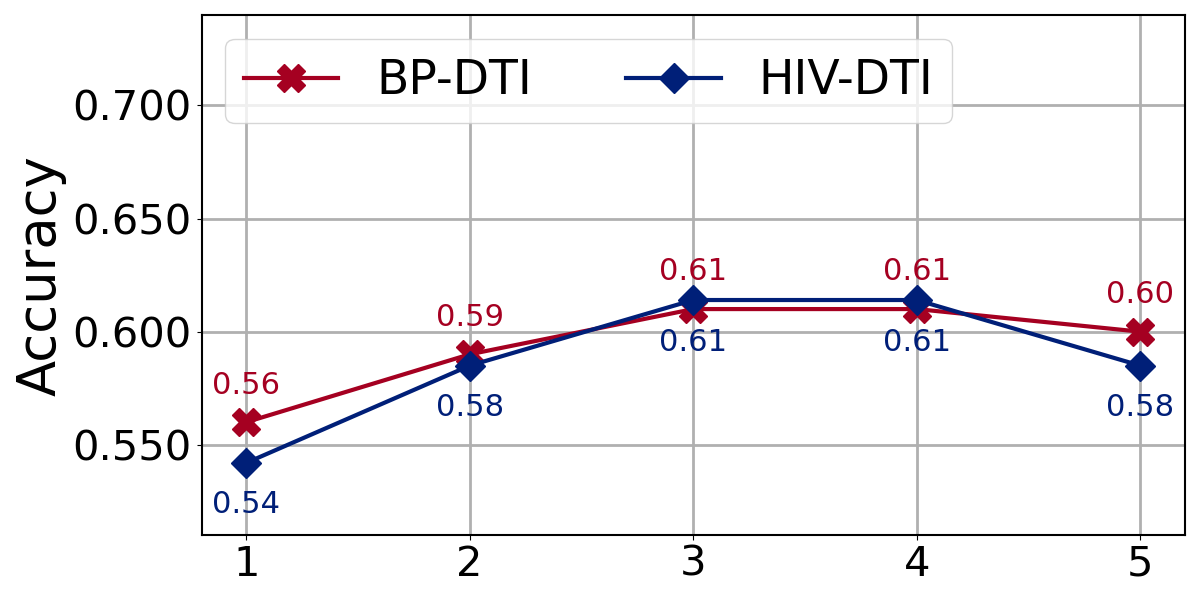}}
    \subfloat[Dimension of Representations]{
    \includegraphics[width=0.32\linewidth]{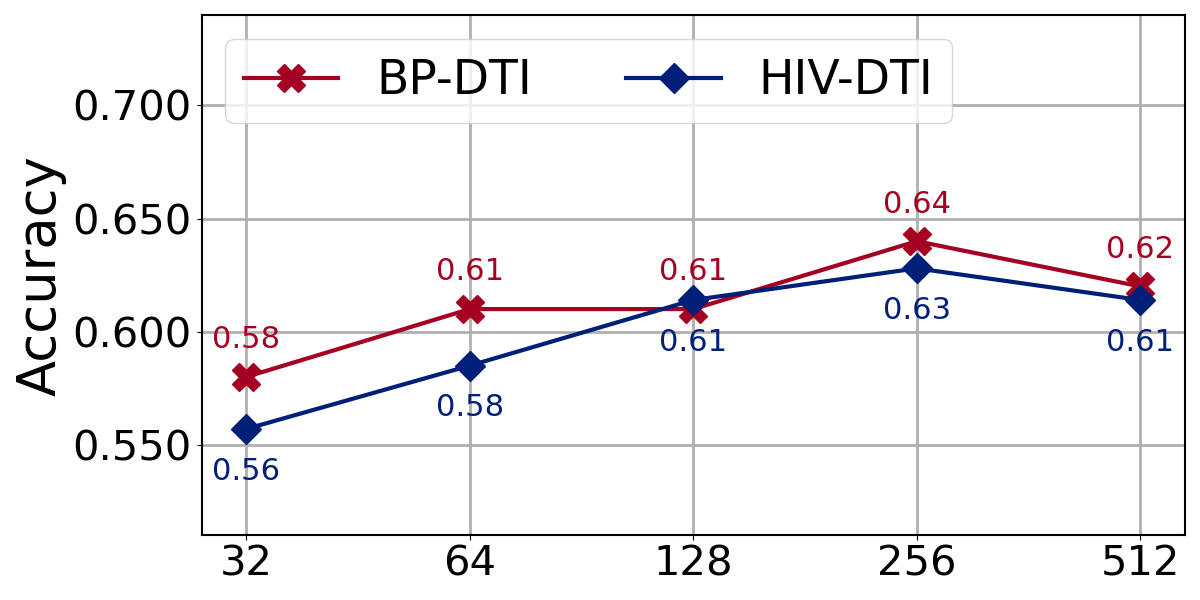}}
    \vspace{-1mm}
    \caption{Hyperparameter sensitivity analysis of BN-GNN on the two datasets.} 
    \vspace{-2mm}
\label{fig_Line}
\end{figure}

\subsection{Hyperparameter Analysis (RQ3)}
\label{Sec_Hyperparameter}
\revised{We also explore the floating perturbations of key hyperparameters in three modules of BN-GNN, namely the neighbor pre-defined amount in the network building module, the action set's size in the meta-policy module, and the dimension of the network representation in the GNN module.} Fig.~\ref{fig_Line}(a) shows that increasing the number of neighbors (determined by k of KNN) when building the adjacency matrix does not always yield better network representations. The possible reason is that each region in the brain network only has meaningful connections with a limited number of neighbors. From Fig.~\ref{fig_Line}(b), we observe that the performance of BN-GNN is often the best when the number of aggregations is 3. When the action space is further expanded, BN-GNN still maintains a relatively stable performance. These two phenomena verify that the best representation of most brain networks can be obtained within 3 aggregations, and BN-GNN is robust to \revised{the fluctuation of action set size (i.e., all possible actions, which is also the maximum aggregation iterations that may occur)}. The results in Fig.~\ref{fig_Line}(c) show that unless the dimension is too small, the performance of BN-GNN will not \revised{fluctuate excessively by updating the embedding dimension.}

\section{Conclusions and Future Work}
\label{Sec_Conclusions}
\revised{In this work, a GNN-based brain network representation framework, namely BN-GNN is proposed.
In particular, BN-GNN combines DRL and GNN for the first time} to achieve customized aggregation for different networks, effectively improving traditional GNNs in brain network representation learning. Experimental results \revised{imply that BN-GNN consistently outperforms state-of-the-art baselines on eight brain network disease analysis tasks.}
In future work, we will improve BN-GNN from both technical and practical aspects.
Technically, we discuss the idea of automatically searching hyperparameters for our model BN-GNN as a future trend. 
Practically, we state the explainability of our model, which is an emerging area in brain network analysis.
Specifically, even if we observe that changes in the number of neighbors have little effect on the performance of BN-GNN, setting this hyperparameter manually is not an optimal solution.
Therefore, multi-agent reinforcement learning can be introduced into BN-GNN to automatically search for the optimal brain network structure and model layer number.
\revised{Besides, for applications of brain disease analysis,} the explainability of the classification results is often as important as the accuracy.
In other words, while successfully predicting a damaged brain network, it is also necessary to understand which regions within the network are responsible for the damage.
Therefore, to improve the explainability, \revised{better pooling or interpretation techniques like} attention-based pooling and grad-cam, can be introduced to BN-GNN.

\section*{Declaration of Competing Interest}
The authors declare that they have no known competing financial interests or personal relationships that could have appeared to influence the work reported in this paper.

\section*{Acknowledgment}
The authors of this paper were supported by the National Key R\&D Program of China through grant 2021YFB1714800, NSFC, China through grants U20B2053 and 62073012, S\&T Program of Hebei, China through grant 20310101D, Beijing Natural Science Foundation, China through grants 4202037 and 4222030. Philip S. Yu was supported by the NSF under grants III-1763325, III-1909323, and SaTC-1930941. We also thank CAAI-Huawei MindSpore Open Fund and Huawei MindSpore platform for providing the computing infrastructure.

\bibliography{reference}{}
\bibliographystyle{elsarticle-num-names}
\end{document}